%% file: main.tex
\documentclass[10pt,twocolumn,letterpaper]{article}

\usepackage{cvpr}              % To produce the CAMERA-READY version
\input{preamble}

\definecolor{cvprblue}{rgb}{0.21,0.49,0.74}
\usepackage[pagebackref,breaklinks,colorlinks,allcolors=magenta]{hyperref}
\usepackage{amssymb,bbding,pifont}
\newcommand{\yeying}[1]{\textcolor{black}{#1}}
\definecolor{myblue}{RGB}{66,133,244}
\definecolor{mygreen}{RGB}{51,168,83}
\definecolor{myyellow}{RGB}{251,188,3}
\definecolor{myred}{RGB}{234,67,53}
\definecolor{mygrey}{RGB}{95,99,104}
\definecolor{mypup}{RGB}{153,0,204}

\def\paperID{*****} % *** Enter the Paper ID here
\def\confName{CVPR}
\def\confYear{2025}
\hypersetup{pagebackref,breaklinks,colorlinks}
\title{NTIRE 2025 Challenge on Day and Night Raindrop Removal for Dual-Focused Images: Methods and Results}

\author{Xin Li$^\dagger$ \quad Yeying Jin$^\dagger$ \quad Xin Jin$^\dagger$ \quad Zongwei Wu$^\dagger$ \quad Bingchen Li$^\dagger$ \quad Yufei Wang$^\dagger$ \quad Wenhan Yang$^\dagger$ \\ 
Yu Li$^\dagger$ \quad Zhibo Chen$^\dagger$ \quad Bihan Wen$^\dagger$ \quad Robby T. Tan$^\dagger$ \quad Radu Timofte$^\dagger$ \quad Qiyu Rong \\ 
Hongyuan Jing \quad Mengmeng Zhang \quad Jinglong Li \quad Xiangyu Lu \quad Yi Ren \quad
Yuting Liu \\ 
Meng Zhang \quad Xiang Chen \quad Qiyuan
Guan \quad Jiangxin Dong \quad Jinshan Pan \quad Conglin Gou \\
Qirui Yang \enspace Fangpu Zhang \enspace Yunlong Lin \enspace Sixiang Chen\enspace Guoxi Huang \enspace Ruirui Lin \enspace Yan Zhang\quad  \\Jingyu Yang \quad Huanjing Yue \quad Jiyuan Chen \quad Qiaosi Yi \quad Hongjun Wang \quad Chenxi Xie \\Shuai Li\quad  Yuhui Wu \quad Kaiyi Ma \quad Jiakui Hu \quad Juncheng Li \quad Liwen Pan \quad Guangwei Gao \\ Wenjie Li  \quad Zhenyu Jin \quad Heng Guo\quad Zhanyu Ma \quad Yubo Wang \quad Jinghua Wang \quad Wangzhi Xing  \\ Anjusree Karnavar \quad Diqi Chen\quad Mohammad Aminul Islam \quad Hao Yang \quad Ruikun Zhang\\ Liyuan Pan \quad Qianhao Luo \quad XinCao \quad Han Zhou \quad Yan Min\quad Wei Dong\quad Jun Chen \\ Taoyi Wu \quad Weijia Dou\quad
Yu Wang\quad Shengjie Zhao \quad Yongcheng Huang \quad Xingyu Han\\Anyan Huang  \quad Hongtao Wu \quad Hong Wang \quad Yefeng Zheng\quad Abhijeet Kumar\\ Aman Kumar \quad  Marcos V. Conde \quad Paula Garrido\quad Daniel Feijoo\quad Juan C. Benito \\ Guanglu Dong \quad Xin Lin \quad Siyuan Liu\quad Tianheng Zheng\quad Jiayu Zhong\quad  Shouyi Wang\\ Xiangtai Li \quad Lanqing Guo\quad  Lu Qi \quad Chao Ren \quad Shuaibo Wang \quad  Shilong Zhang\\  Wanyu Zhou\quad  Yunze Wu\quad  Qinzhong Tan \quad Jieyuan Pei\quad Zhuoxuan Li \quad Jiayu Wang \\ Haoyu Bian \quad Haoran Sun \quad Subhajit Paul \quad Ni Tang \quad Junhao Huang\quad Zihan
Cheng\\ Hongyun Zhu\quad Yuehan Wu \quad Kaixin Deng \quad 
Hang Ouyang\quad Tianxin Xiao\quad Fan Yang\\ Zhizun Luo \quad Zeyu Xiao\quad Zhuoyuan Li \quad Nguyen Pham Hoang Le \quad An Dinh Thien \\ Son T. Luu \quad Kiet Van Nguyen \quad Ronghua Xu \quad Xianmin Tian\quad Weijian Zhou\\ Jiacheng Zhang \quad Yuqian Chen \quad Yihang Duan \quad Yujie Wu \quad Suresh Raikwar \quad Arsh Garg \\ Kritika \quad Jianhua Zheng \quad Xiaoshan Ma\quad Ruolin Zhao\quad Yongyu Yang\quad Yongsheng Liang\\ Guiming Huang \quad Qiang Li  \quad  Hongbin Zhang\quad Xiangyu Zheng \quad A.N. Rajagopalan}

\begin{document}
\maketitle
\renewcommand{\thefootnote}{}
\footnotetext{$^{\dagger}$X. Li, Y. Jin, X. Jin, Z. Wu, B. Li, Y. Wang, W. Yang, Y. Li, Z. Chen, B. Wen, R. Tan and R. Timofte are the challenge organizers. The contact details can be found in section~\ref{organizers}. (Corresponding authors: Xin Li, Yeying Jin and Xin Jin).}
\footnotetext{The other authors are participants of the NTIRE 2025 Challenge on Day and Night Raindrop Removal for Dual-Focused
Images.}
\footnotetext{The NTIRE2025 website:~\url{https://cvlai.net/ntire/2025/}}
\footnotetext{The Competition website~\url{https://codalab.lisn.upsaclay.fr/competitions/21345}}
\footnotetext{The Raindrop Clarity database:~\url{https://github.com/jinyeying/RaindropClarity}}

\input{sec/0_abstract}    
\input{sec/1_intro}

\input{sec/2_Challenge}
\input{sec/3_Challenge_Results}
\input{sec/4_Teams_and_Methods_supp}

\input{sec/5_Appendix}

{
    \small
    \bibliographystyle{ieeenat_fullname}
    \bibliography{main}
}

% WARNING: do not forget to delete the supplementary pages from your submission 
% \input{sec/X_suppl}

\end{document}

%% file: preamble.tex
\usepackage[dvipsnames]{xcolor}

%% file: sec/0_abstract.tex
\begin{abstract}
This paper reviews the NTIRE 2025 Challenge on Day and Night Raindrop Removal for Dual-Focused Images. This challenge received a wide range of impressive solutions, which are developed and evaluated using our collected real-world Raindrop Clarity dataset~\cite{jin2024raindrop}. Unlike existing deraining datasets, our Raindrop Clarity dataset is more diverse and challenging in degradation types and contents, which includes \yeying{day raindrop-focused, day background-focused, night raindrop-focused, and night background-focused} degradations. This dataset is divided into three subsets for competition: \yeying{14,139} images for training, 240 images for validation, and 731 images for testing. The primary objective of this challenge is to establish a new and powerful benchmark for the task of removing raindrops under varying lighting and focus conditions. There are a total of 361 participants in the competition, and 32 teams submitting valid solutions and fact sheets for the final testing phase. These submissions achieved state-of-the-art (SOTA) performance on the Raindrop Clarity dataset. The project can be found at~\url{https://lixinustc.github.io/CVPR-NTIRE2025-RainDrop-Competition.github.io/}.
\end{abstract}

%% file: sec/1_intro.tex
\section{Introduction}
\label{sec:intro}
% Image deraining has attracted lots of 

%% cross-referencing NTIRE 2025 associated challenges
Image deraining has been a long-standing research topic in low-level image processing~\cite{wang2019spatialSPAData,zamir2022restormer,tu2022maxim,wu2024rainmamba,valanarasu2022transweather,li2023learningDIL,li2020learningCNNderain,zou2024freqmambaDerain,liu2020lira,li2025hybridagents}, aiming to remove visual artifacts caused by rain streaks~\cite{yang2017deep-rain100} or raindrops~\cite{qian2018attentiveRainDrop} under adverse weather conditions. It plays a crucial role not only in enhancing the perceptual quality of images but also in improving the performance of high-level vision tasks, such as autonomous driving and pedestrian detection. 
With the rapid development of diverse backbone architectures in low-level image processing~\cite{wang2019spatialSPAData,zamir2022restormer,tu2022maxim,wu2024rainmamba,valanarasu2022transweather,li2023learningDIL,pang2020fan,li2020learningCNNderain,zou2024freqmambaDerain,li2024sed,liang2021swinir,li2022hst}, several deraining benchmarks have been introduced, incorporating architectures such as ResNet~\cite{zhang2018densityDID-Data, wang2019spatialSPAData, li2020learningCNNderain}, Transformer~\cite{xiao2022imageTransformerDerain, chen2024rethinkingTransformerDerain}, MLP~\cite{tu2022maxim}, and Mamba-based frameworks~\cite{zou2024freqmambaDerain}. 
Moving beyond single-task benchmarks, recent studies have begun to explore multi-task deraining methods~\cite{valanarasu2022transweather,ozdenizci2023restoringweatherdiff} or unified frameworks capable of handling various adverse weather conditions simultaneously, such as rain, haze and snow removal. These multi-task deraining methods impose new challenge to the model design, which requires higher adaptability and generalization capabilities. Furthermore, to enhance subjective visual quality~\cite{yu2024sfiqa,zhang2018lpips,lu2024aigcvqa}, diffusion-based deraining methods~\cite{li2023diffusionsurvey,huang2024revivediff,chen2024teachingt3diffusion,ozdenizci2023restoringweatherdiff,shen2023rethinkingRainDiff}, such as WeatherDiff~\cite{ozdenizci2023restoringweatherdiff}, have recently emerged, demonstrating promising results and opening new opportunities for advanced research on image deraining.

Datasets are essential for evaluating the effectiveness of deraining algorithms. In the early stages, rain degradation was typically \yeying{synthesized~\cite{hao2019learningRaindropdataset,li2016rainRain12,yang2017deepRain100HL,fu2017removingDDN-data,zhang2018densityDID-Data,li2019singleRain800,hu2019depthRainCityScapes,li2019heavyOutdoor-Rain,jiang2020multiRain13K}}, as capturing paired rainy and clean images simultaneously is extremely challenging, even with professional cameras and controlled environments. To address this limitation, recent works have proposed methods that generate real-world rainy degradation by utilizing the video temporal priors~\cite{wang2019spatialSPAData} or inserting the glass with adherent \yeying{raindrops~\cite{qian2018attentiveRaindropdataset,soboleva2021raindropsRaindropdataset,porav2019canRaindropdataset} and sprayed water~\cite{quan2021removingRainDS}}. Raindrops, as the typical rain degradation type, usually appear on camera lenses or windshields, which significantly reduces image visibility in human life, posing challenges for applications like surveillance and autonomous driving. Effective raindrop removal is crucial to ensuring reliable performance in these systems. However, most deraining datasets overlooked the complex environment in real life. In particular, there are few raindrop-focused datasets, and most existing datasets focus on capturing background scenes while the camera is focused on the background. Meanwhile, these datasets and related works primarily target daytime scenarios, with limited attention to nighttime conditions~\cite{lin2024nightrain,jin2022unsupervised,jin2023enhancing,lin2024nighthaze}.

To address the aforementioned challenges, we collaborate with the 2025 New Trends in Image Restoration and Enhancement (NTIRE \yeying{2025}) workshop to organize the first Challenge on Day and Night Raindrop Removal for Dual-Focused Images. The challenge is based on our Raindrop Clarity Dataset~\cite{jin2024raindrop}, which contains 5,442 daytime and 4,712 nighttime raindrop image pairs or triplets. Its primary objective is to promote research on real-world rain removal under varying lighting conditions while simultaneously restoring fine image textures degraded by raindrop-induced defocus. This competition attracted a total of 361 participants, resulting in 32 teams submitting their methods and results, which are documented in this challenge report. We believe that this challenge will encourage further advancements of research on image deraining.

This challenge is one of the NTIRE 2025~\footnote{\url{https://www.cvlai.net/ntire/2025/}} Workshop associated challenges on: ambient lighting normalization~\cite{ntire2025ambient}, reflection removal in the wild~\cite{ntire2025reflection}, shadow removal~\cite{ntire2025shadow}, event-based image deblurring~\cite{ntire2025event}, image denoising~\cite{ntire2025denoising}, XGC quality assessment~\cite{ntire2025xgc}, UGC video enhancement~\cite{ntire2025ugc}, night photography rendering~\cite{ntire2025night}, image super-resolution (x4)~\cite{ntire2025srx4}, real-world face restoration~\cite{ntire2025face}, efficient super-resolution~\cite{ntire2025esr}, HR depth estimation~\cite{ntire2025hrdepth}, efficient burst HDR and restoration~\cite{ntire2025ebhdr}, cross-domain few-shot object detection~\cite{ntire2025cross}, short-form UGC video quality assessment and enhancement~\cite{ntire2025shortugc,ntire2025shortugc_data}, text to image generation model quality assessment~\cite{ntire2025text}, day and night raindrop removal for dual-focused images~\cite{ntire2025day}, video quality assessment for video conferencing~\cite{ntire2025vqe}, low light image enhancement~\cite{ntire2025lowlight}, light field super-resolution~\cite{ntire2025lightfield}, restore any image model (RAIM) in the wild~\cite{ntire2025raim}, raw restoration and super-resolution~\cite{ntire2025raw} and raw reconstruction from RGB on smartphones~\cite{ntire2025rawrgb}.

%% file: sec/2_Challenge.tex
\section{Challenge}
\label{sec:challenge}
The NTIRE 2025 Challenge on Day and Night Raindrop Removal for Dual-Focused Images is the first competition to be organized to advance the development of real-world image draining under different light conditions and focusing degrees. The details of the whole challenge are as follows:
\subsection{Datasets}
The dataset used in this challenge is the RainDrop Clarity dataset~\cite{jin2024raindrop}, which includes daytime and nighttime scenes for training, validation, and testing. The original dataset includes both daytime and nighttime scenes for training and testing: (i) the daytime raindrop dataset contains a total of 5,442 paired or triplet images, with 4,713 pairs/triplets used for training and the remaining 729 pairs/triplets for testing; \yeying{Specifically, among the 4,713 daytime training image pairs/triplets, 1,575 are background-focused while 3,138 are raindrop-focused;} (ii) the nighttime raindrop dataset consists of 9,744 paired or triplet images, where 8,655 pairs/triplets are allocated to the training set, and the remaining 1,089 pairs/triplets are reserved for the validation and test set. \yeying{Specifically, among the 8,655 nighttime training image pairs/triplets, 4,143 are background-focused while 4,512 are raindrop-focused.}

In this challenge, the Raindrop Clarity dataset is divided into training, validation, and testing subsets, where the training set comprises \yeying{4,713 triplets, totaling 14,139 images}, while the validation and testing sets contain 240 and 731 image, respectively. To ensure diversity and prevent distribution bias, the validation and testing sets are reorganized to maintain a balanced distribution in various rainy scenes. In particular, during the validation and testing phases, intermediate rain-free blurry images are withheld to better simulate real-world application scenarios. Additionally, to rigorously evaluate the robustness of the algorithms submitted, no explicit distinction is made between day- and night-time scenes in the validation and testing subsets. 
\yeying{The training set for this challenge is exactly the same as that used in the Raindrop Clarity dataset paper~\cite{jin2024raindrop}. The validation set of 240 images consists of 120 daytime and 120 nighttime images, each including 60 raindrop-focused and 60 background-focused samples. The test set of 731 images contains 381 daytime images and 350 nighttime images. Among the 381 daytime images, 294 are raindrop-focused and 87 are background-focused. For the 350 nighttime images, 230 are raindrop-focused and 120 are background-focused.} 

% For this challenge, the Raindrop Clarity dataset is divided into three parts, including training, validation, and testing parts, where validation and testing parts are processed to keep the balance between different rainy scenes, resulting in 13368 triplet images for training, 240 pairs of images for validation, and 731 pairs of images for testing. In the validation and testing stages, the middle rainy-free blur images are not provided to satisfy the real-world application. Moreover, to validate the robustness of the algorithms from participants, we do not distinguish the day/night scenes in the validation/testing phases. 

\subsection{Evaluation Protocol}
This challenge utilizes three metrics to measure the objective and subjective quality of restored images, \ie, PSNR, SSIM, and LPIPS, respectively. The final score used for ranking is computed by reweighting the above metrics as:
\begin{equation}
    \mathrm{Score} = 10\times\mathrm{PSNR (Y)} + 10\times \mathrm{SSIM (Y)} - 5\times \mathrm{LPIPS},
    \label{eq:finalscore}
\end{equation}
where $(\mathrm{Y})$ denotes the PSNR and SSIM are measured with the $\mathrm{Y}$ channel after coverting the image from RGB space to the $\mathrm{YCbCr}$ space. For $\mathrm{LPIPS}$, we first normalize the image pixel values into the range $[-1, 1]$, then utilize the Alex network configuration for distance measurement between restored images and ground-truth images.
PSNR and SSIM, two widely used metrics in deraining~\cite{chen2024dual,chen2023sparse,chen2024teaching,ye2023adverse,jarvisir2025} and restoration tasks~\cite{jin2021dc,jin2022structure,jin2023estimating,jin2024des3}.

\subsection{Challenge Phases}
There are two phases in this challenge, \ie, the development and testing phases. The details are as follows.

\noindent\textbf{Development Phase:}
In the development phase, we release \yeying{4,713 triplets, totaling 14,139 images} in our Raindrop Clarity dataset, including daytime and nighttime rainy images, rain-free blur images, and their corresponding ground-truth images, to support each team in developing their algorithms. Moreover, we release 240 daytime and nighttime rainy images without their ground truth for validation. Each participant can upload their restored images to the challenge platform by removing the raindrop with their developed algorithm. Then they can obtain the corresponding final score, PSNR, SSIM, and LPIPS. In the development phases, we received 1264 submissions from 74 teams in total. 

\begin{table*}[tp]
    \centering
    \caption{Quantitative results of the NTIRE 2025 on Day and Night Raindrop Removal for Dual-Focused Images. The best and second results are in \textcolor{red}{red} and  \textcolor{blue}{blue}, respectively.}
    \resizebox{\textwidth}{!}{\begin{tabular}{cc|cccc|cc|c|c|c}
    \toprule
       Teams & Leader & Final Score $\uparrow$ & PSNR$\uparrow$ & SSIM$\uparrow$ & LPIPS $\downarrow$ & Params. (M) & GFlops (G) & Ensemble & Extra Data  &  Rank  \\ \midrule
       Miracle & Qiyu Rong & \textcolor{red}{34.3518} &\textcolor{blue}{27.6619} & \textcolor{red}{0.8087} &\textcolor{blue}{0.2794} & 26.89 & 42.33 & $\otimes$ & $\otimes$  &  1
       \\
       EntroVision & Xiang Chen &\textcolor{blue}{34.2145} &\textcolor{red}{27.6629} &	\textcolor{blue}{0.8071} &	0.3038& 16.6 & 129.9 & $\checkmark$ & $\checkmark$ &2
       \\
IIRLab & Conglin Gou & 33.4940 &	26.9564 &	0.7993 &	0.2910 & 11.69 & \textcolor{red}{12.62} & $\checkmark$ & $\otimes$  & 3
       \\
      PolyRain & Jiyuan Chen & 32.7910 &	26.1945 & 0.7701 &\textcolor{red}{0.2210} & 26 & 183 & 
       
       $\checkmark$ & $\otimes$ & 4 \\
       H3FC2Z & Kaiyi Ma & 32.5652 &	26.3954 &	0.7737&	0.3133 &  36.2 & 120 & $\checkmark$ & $\otimes$ & 5  \\
       IIC Lab & Juncheng Li & 32.5367 &	26.1606&	0.7517&	0.2281& 45.5 & 230 & $\otimes$ & $\otimes$ & 6 \\
       BUPT CAT & Wenjie Li & 32.4091 & 26.2216 &	0.7774 & 0.3172 & 46.1 & 139 & $\checkmark$ & $\otimes$ & 7 \\
       WIRTeam & Yubo Wang & 32.2750 &	26.0379 &	0.7714 & 0.2954 & 35.84 & 265.88 & $\otimes$ & $\otimes$ & 8 \\ 
       GURain & Wangzhi Xing & 32.1223 & 25.9473 & 0.7671 & 0.2993 & 25.31 & 87.7 &  $\otimes$ & $\otimes$ & 9 \\
       BIT\_ssvgg & Hao Yang & 31.9508 &	25.7955 &	0.7665 &	0.3019&68 & 742& $\otimes$ & $\otimes$ & 10 \\
       CisdiInfo-MFDehazNet & Qianhao Luo & 31.9346  &	25.7951  &	0.7670  &	0.3062 & 5 & 132 & $\otimes$ & $\otimes$ & 11 \\
       McMaster-CV & Han Zhou & 31.8097 & 25.7445 & 0.7533 & 0.2936 & 11.15 & 158.32 & $\otimes$ & $\otimes$ & 12 \\
       Falconi & Taoyi Wu & 31.6836 & 	25.4967 & 	0.7506 & 	0.2638 & 109.59 & 522 & $\otimes$ & $\otimes$ & 13 \\
       Dfusion & Yongcheng Huang & 31.6098 &	25.5081 &	0.7469 &0.2735 & - & 105 & $\otimes$ & $\otimes$ & 14 \\
        RainMamba & Hongtao Wu & 31.4608 &	25.4251 &	0.7509 &	0.2946 & 34.49 & 119 & $\otimes$ & $\otimes$ & 15 \\
        RainDropX & AbhijeetKumar & 31.1777 &	25.1864 &	0.7518 &	0.3053& 26.1 & 141 &  $\otimes$ & $\otimes$  & 16 \\
       Cidaut AI & Marcos V. Conde & 31.1667 &	25.3108 &	0.7456 &	0.3201 & \textcolor{myred}{2.38} & 64.28  & $\otimes$ & $\otimes$ & 17 \\
       DGL\_DeRainDrop & Guanglu Dong & 31.0267 &	24.9668 &	0.7397 & 0.2674 & 129.99 &  91.03 & $\otimes$ & $\otimes$  & 18 \\
       xdu\_720 & Shuaibo Wang& 31.0051&	25.1503&	0.7535& 0.3360& 148 & 205 & $\otimes$ & $\otimes$ & 19 \\
       EdgeClear-DNSST Team &Jieyuan Pei &  30.8227 & 25.0620 &0.7384 & 0.3246 & 23 & 175 & $\otimes$ & $\otimes$ & 20 \\
       MPLNet & Jiayu Wang & 30.7730 &	24.6601 &	0.7178 & 0.2130 & 10.7 & 82.3296 & $\otimes$ & $\otimes$ & 21 \\
       Singularity & Subhajit Paul & 30.4006 &  24.4232 &    0.7256  &  0.2556 & 46 & 760 & $\otimes$ & $\otimes$ & 22 \\
       VIPLAB & Ni Tang & 30.3678 &	24.3905 &	0.7203 &	0.2452 & 5.97 & 147 & $\otimes$ & $\otimes$ & 23 \\
       2077Agent & Kaixin Deng & 30.3581 & 	24.5354 & 	0.7209 & 	0.2773& 28 & - & $\otimes$ & $\otimes$ & 24 \\
       X-L & Zeyu Xiao & 30.2202&	24.5935&	0.7474&	0.3694&12.63& 120 & $\checkmark$ & $\otimes$ & 25 \\
       UIT-SHANKS & Nguyen Pham Hoang Le & 30.0521 & 	24.5982 & 	0.7361 & 	0.3814&  - & - &  $\checkmark$ & $\otimes$ & 26 \\
       One Go Go & Ronghua Xu & 30.0126 & 	24.1069 & 	0.7241& 	0.2671& 25.31 & \textcolor{myblue}{33} & $\checkmark$ & $\otimes$ & 27 \\
DualBranchDerainNet & Yuqian Chen & 29.7762& 	24.6438& 	0.7247 & 0.4229&  \textcolor{myblue}{2.5} & - & $\otimes$ & $\otimes$ & 28 \\
QWE & Yihang Duan & 29.5937 &	24.2767 &	0.6987 &0.3340& - & -& $\otimes$ & $\otimes$ & 29 \\
Visual and Signal Information Processing Team  & Suresh Raikwar & 24.5878 &	21.8319 &	0.6018 &	0.6523&- & - & $\otimes$ & $\otimes$ & 30 \\ 
The Zheng family group & Jianhua Zheng & 23.9370 &	21.0757 &	0.5524 &	0.5326 & -& -& -& - & 31 \\
 RainClear Pioneers& Qiang Li &  22.7072 & 18.3433 &  0.6521&  0.4314 & - & - & - & - & 32 \\
       \bottomrule
    \end{tabular}}
    \label{tab:results}
\end{table*}

\noindent\textbf{Testing Phases:}
In the testing phases, we release 731 daytime and nighttime rainy images without their ground truth for testing. To ensure the fairness of this challenge, we hide the leaderboard, where each team cannot find the performance of other teams. The final ranking is achieved with the score in Eq.~\ref{eq:finalscore}. In the test stage, 75 teams submitted their final results to the challenge platform. At the end of this competition, we received the fact sheets and source codes from 32 teams, which are utilized for final ranking. 

%% file: sec/3_Challenge_Results.tex
\section{Challenge Results}
We have summarized the challenge results in Table~\ref{tab:results}. 
\yeying{Among all submissions, Team Miracle ranked 1st with the highest final score of 34.3518, achieving the top performance across PSNR (27.6619) and SSIM (0.8087), as well as a competitive LPIPS score (0.2794), despite using a relatively moderate parameter count (26.89M) and no ensemble or extra data.
The 2nd place, EntroVision, delivered a final score of 34.2145. It achieved high PSNR (27.6629) and SSIM (0.8071) values, along with a relatively low LPIPS (0.3038). The model benefited from the use of ensemble and extra data.
The 3rd place, IIRLab, achieved comparable accuracy with a final score of 33.4940, PSNR of 26.9564, SSIM of 0.7993, and LPIPS of 0.2910. Notably, it accomplished this while maintaining a lightweight design—using only 11.69M parameters and 12.62GFlops, despite employing an ensemble strategy, demonstrating an excellent trade-off between accuracy and efficiency.
Overall, the table reveals a wide diversity in model complexity, ranging from lightweight solutions to highly complex networks, reflecting different trade-offs between performance and resource consumption.}

%% file: sec/4_Teams_and_Methods_supp.tex
% \clearpage
% \setcounter{page}{1}
% \maketitlesupplementary

\section{Teams and Methods}
\label{sec:teams_and_methods}
\subsection{Miracle}
% revised by Miracle, please finish it before 12/04/2025.
\begin{figure}[htbp]
    \centering
    \includegraphics[width=1.0\linewidth]{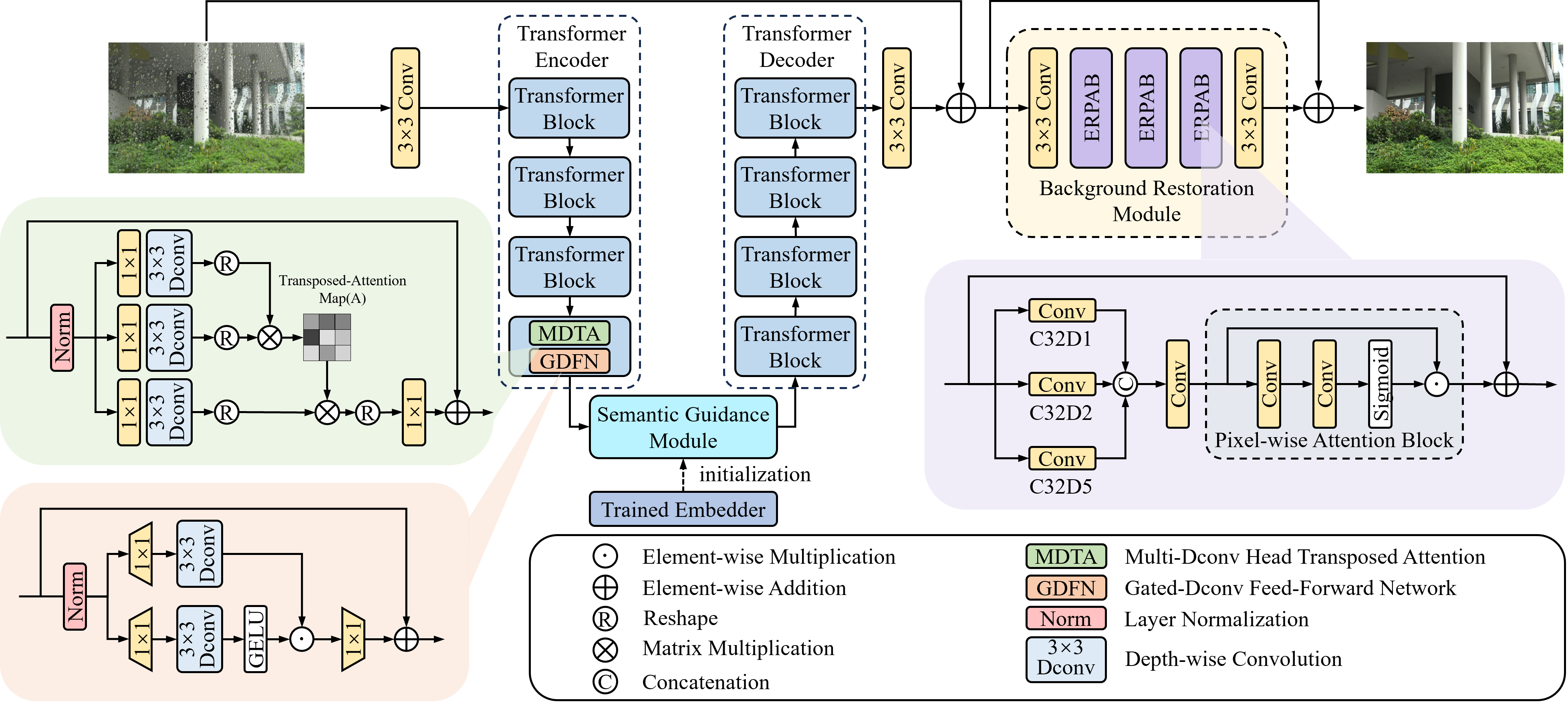}
    \caption{The framework of STRRNet, proposed by Team Miracle}
    \label{fig:Miracle_1}
\end{figure}

\begin{figure}[htbp]
    \centering
    \includegraphics[width=1.0\linewidth]{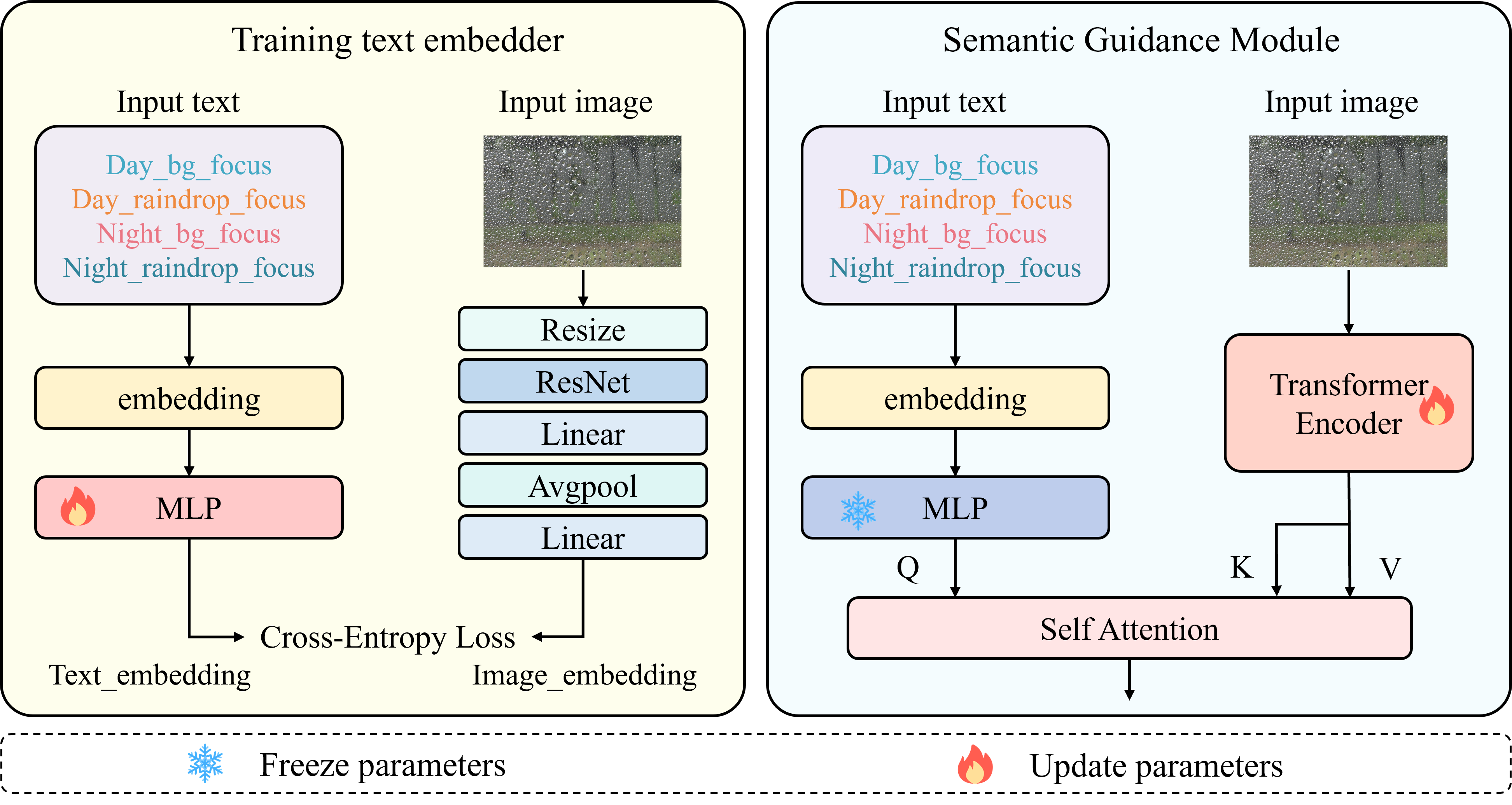}
    \caption{Semantic Guidance Module of STRRNet, proposed by Team Miracle}
    \label{fig:Miracle_3}
\end{figure}

\begin{figure}[htbp]
    \centering
    \includegraphics[width=1.0\linewidth]{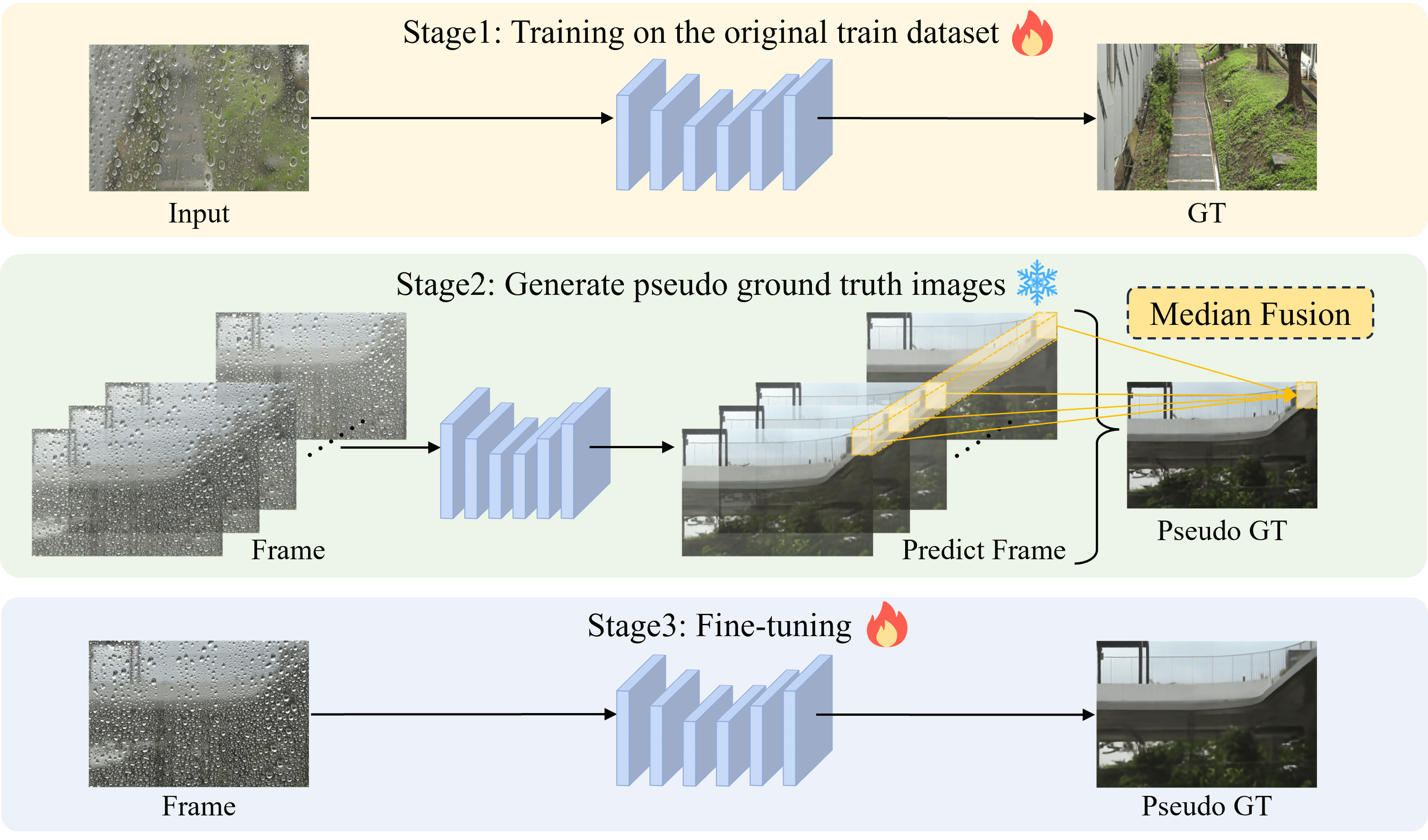}
    \caption{Training Strategy of STRRNet, proposed by Team Miracle}
    \label{fig:Miracle_2}
\end{figure}

This team proposes the STRRNet~\cite{rong2025STRRNet}, which is developed based on Restormer \cite{zamir2022restormer}, as shown in Fig~\ref{fig:Miracle_1}. They categorize the training images into four classes based on lighting conditions and raindrop types: night\_bg\_focus, night\_raindrop\_focus,
day\_bg\_focus, and day\_raindrop\_focus. As shown in Fig~\ref{fig:Miracle_3}, a text embedder is first trained on the labeled training set using these four categories. Then, they design a semantic
guidance module, which is added at the end of the Restormer encoder.
This module utilizes the encoded image features from Restormer to guide the decoder in performing distinct image restoration operations for the
four different types of images. Additionally, they introduce a background restoration
subnetwork at the output of Restormer, which consists of multiple
convolutional layers to enhance image details. 
The training strategy of STRRNet is illustrated in the Fig \ref{fig:Miracle_2}.
First, they train a pre-trained model on the original training dataset. This model is then used to perform inference on all frames within the same scene in the test set. The inferred images may still contain residual raindrops and artifacts. Since the background remains consistent across different time frames while raindrop positions vary temporally, this motivates them to perform median fusion across multiple frames from the same scene to obtain a median-fused image. Due to the dynamic nature of raindrop artifacts, their inconsistent locations across frames typically prevent them from appearing in the median values, whereas the stable background information is preserved. They then treat the median-fused image as a pseudo ground truth and use it in a semi-supervised fine-tuning phase to enhance the model’s raindrop removal capability on unlabeled images.
\\
\textbf{Training description}
The Adam optimizer is used for training, with a total of 500,000 iterations.
The learning rate is set to 0.0003 for the first 9,2000 iterations and then
gradually decays from 0.0003 to 0.000001 for the remaining iterations. Images
are randomly cropped to a fixed size of 128×128 for network training,
and geometric image augmentation is applied. The network is optimized
using the L1 loss function and the multi-scale SSIM loss function, with
weights of 1 and 0.2, respectively. All their experiments were conducted on
an RTX 4090.\\
\textbf{Testing description}
Use a sliding window to move across the image, applying the model for
rain removal on each window. Set the sliding window size to 128×128
with an overlap of 32. Then, obtain a median image by applying median
fusion to the images of the same scene. Finally, perform a weighted sum
of the median image and the original image to obtain the final output.\\

\subsection{EntroVision}
\begin{figure}[htbp]
\centering
\includegraphics[width=1.0\linewidth]{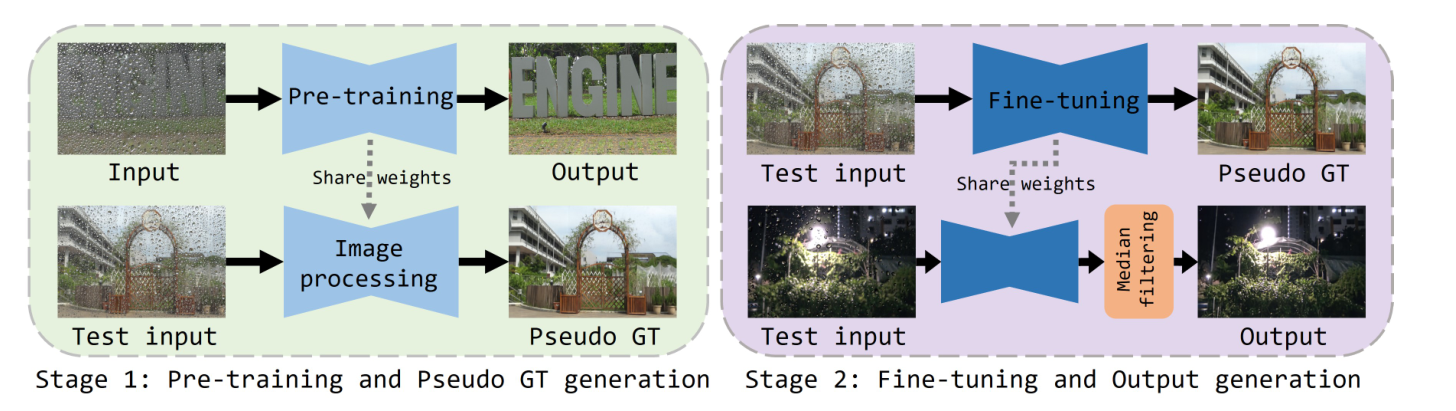}
\caption{Overview of the technique proposed by Team EntroVision for raindrop removal.}
\label{fig:EntroVision}
\end{figure}

This team utilizes a two-stage approach to achieve raindrop removal in this challenge. The technical overview is shown in the Fig.~\ref{fig:EntroVision}. Given the impact of multi-scale features in image deraining~\cite{chen2024bidirectional,chen2023towards}, they pre-train the deraining model in the first stage using MSDT~\cite{chen2024rethinkingMSDT} on the RainDrop Clarity~\cite{jin2024raindrop} training set. To enhance the model's generalization capability, additional pre-training is conducted on the UAV-Rain1k~\cite{chang2024uavrain1k} dataset. Then, the pre-trained deraining model is utilized to obtain the pseudo-ground-truth images for testing images for the test-time learning of the second stage.

In the second stage of the process, they performed fine-tuning using the test samples and the generated paired pseudo ground-truth images. This approach provides a clear direction for transferring pretrained knowledge, rather than simply relying on the model's generalization ability. Subsequently, they process the testing inputs using this fine-tuned deraining model to obtain the final output results. Notably, they designed the deraining network specifically according to the characteristics of the dataset used. To address scenarios where blurry and clear backgrounds might coexist, they employ median filtering to preserve edge information and avoid excessive blurring. Through the above two-stage processing strategies and the specially designed median
filtering technique, they obtain clear deraining results.

\subsection{IIRLab}
\begin{figure}[htbp]
    \centering
    \includegraphics[width=1.0\linewidth]{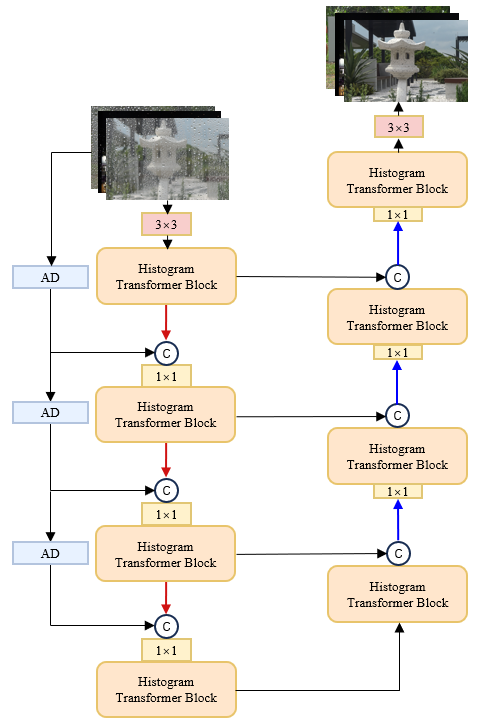}
    \caption{The pipeline of the method proposed by Team IIRLab}
    \label{fig:IIRLab}
\end{figure}

This team observed that the raindrop dataset used in this challenge differs significantly from existing raindrop removal datasets. Traditional datasets primarily focus on background clarity during image shooting, resulting in clean backgrounds and blurry raindrops in the foreground. In such cases, simply removing the raindrops is sufficient to recover a clear background. However, the dataset in this challenge includes a notable portion of images where the camera is focused on the raindrops during shooting, leading to clean raindrops and blurred backgrounds. This introduces a new challenge: removing not only the raindrops themselves but also mitigating background blur caused by defocus, to ultimately recover a clean draining image.

Based on this new question, this team has chosen the Histoformer~\cite{sun2024restoringhistogram} network as their raindrop removal model, as shown in Fig~\ref{fig:IIRLab}. Histoformer is a transformer-based model designed to restore images degraded by severe weather conditions. It incorporates a histogram self-attention mechanism, which sorts and segments spatial features into intensity-based bins and applies self-attention within each bin. This enables the model to focus on spatial features across dynamic intensity ranges and handle long-range dependencies between similarly degraded pixels. Built upon Restormer~\cite{zamir2022restormer}, Histoformer is well-suited for addressing both raindrop artifacts and defocus blur, making it a strong candidate for this task. Based on its architecture and prior performance, the team chose Histoformer as the core model for this challenge.

\noindent\textbf{Training Details.}
This team utilized all image pairs provided in the training set, consisting of raindrop-degraded images (Drop) and their corresponding clean background images (Clean). From this, they extracted 1,200 image pairs for validation, including 400 daytime and 800 nighttime samples. For training, they employed a two-stage training strategy that combines regular training with subsequent fine-tuning. In the first stage, the draining model was trained for 300,000 iterations using the default Histoformer configuration. In the second stage, the model was fine-tuned for an additional 13,000 iterations using only the $\mathcal{L}_1$ loss function. This progressive training approach effectively enhanced model performance, leading to improved final results.

\noindent\textbf{Implememtation Details.}
The implementation is based on PyTorch and was conducted on an NVIDIA RTX 3090 GPU. The network was trained for a total of 300,000 iterations, with an initial batch size of 6 and a patch size of $128\times 128$, following a progressive learning strategy. The team employed the AdamW optimizer with an initial learning rate of $3 \times 10^{-4}$ for the first 92,000 iterations, which was then gradually reduced to $1 \times 10^{-6}$ using a cosine annealing schedule over the remaining 208,000 iterations.
The number of blocks at each stage was set as $L_{i \in {1,2,3,4}} = {4, 4, 6, 8}$, and the channel dimension was fixed at $C = 36$. The channel expansion factor in the DGFF module was set to $r = 2.667$. The number of self-attention heads at each stage was configured as ${1, 2, 4, 8}$, respectively. For data augmentation, horizontal and vertical flips were applied randomly during training.

\subsection{PolyRain}
\begin{figure}[htbp]
    \centering
    \includegraphics[width=1.0\linewidth]{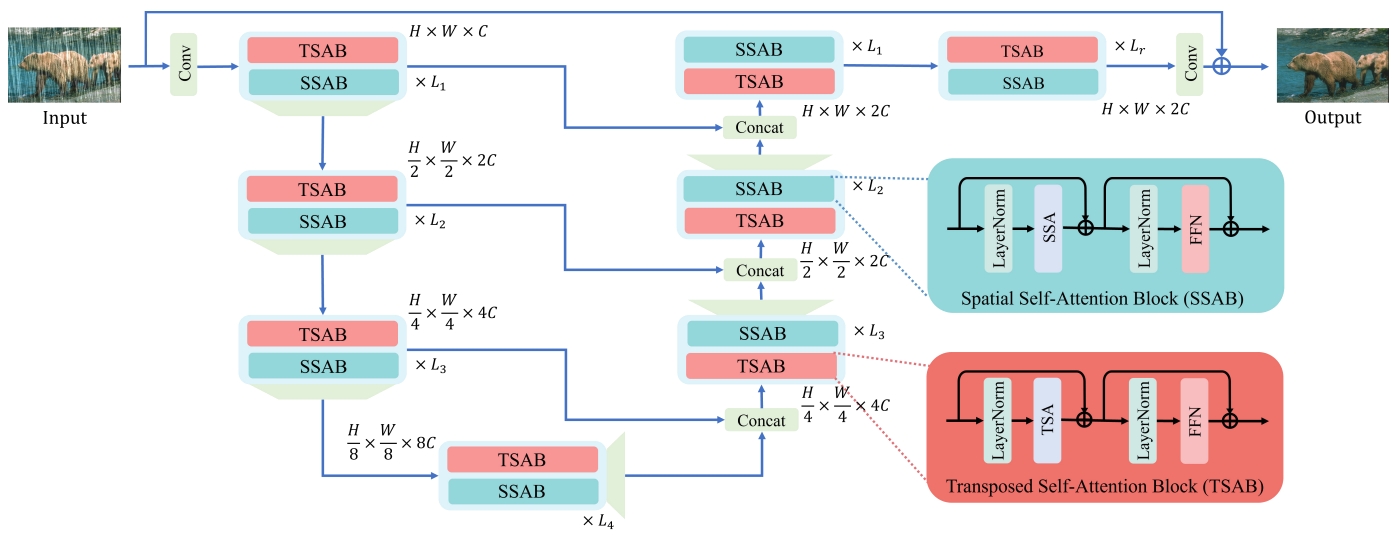}
    \caption{\yeying{The pipeline of the method proposed by Team PolyRain}}
    \label{fig:PolyRain}
\end{figure}

This team initialized and trained a dense X-Restormer model with the given dataset based on Restormer~\cite{zamir2022restormer}. After the first training stage, they finetuned the model on a larger patch size with different loss functions. The whole framework of this team is shown in Fig.~\ref{fig:PolyRain}.

\noindent\textbf{Training Details.}
The training process consists of two phases. In the first phase, the patch size of the training image is set to 256, with a batch size of 8, and a total of 3$\times$105 training iterations. The learning rate is set to $3e^{-4}$ and $\mathcal{L}_1$ Loss is used as the loss function. In the second phase, the model is fine-tuned with a $5e^{-5}$ learning rate and the image patch with 448 $\times$ 448. During this phase, the model is trained simultaneously using $\mathcal{L}_2$ Loss, LPIPS Loss, and SSIM Loss with weights of 1, 0.1, and 0.1, respectively, for 5$\times$104 iterations.

\noindent\textbf{Testing Details.}
To enhance the robustness of the model, the self-ensemble technique is employed. The implementation references the BasicSR library~\footnote{https://github.com/XPixelGroup/BasicSR}. 

\noindent\textbf{Implementation Details.}
This method is implemented based on the famous BasicSR framework~\cite{chan2021basicvsr,chen2024comparative} written in Python. They utilized the AdamW optimizer with an initial learning rate of $3e^{-4}$. Eight A100 GPUs were used for the model training, lasting for about 72 hours for 300000 iterations. In addition, the CosineAnnealingRestart-CyclicLR scheduler was chosen to restart the learning rate at a setting of $[92000, 208000]$. They did not use any efficient optimization strategy or extra datasets.

\subsection{H3FC2Z}
\begin{figure}[htbp]
    \centering
    \includegraphics[width=1.0\linewidth]{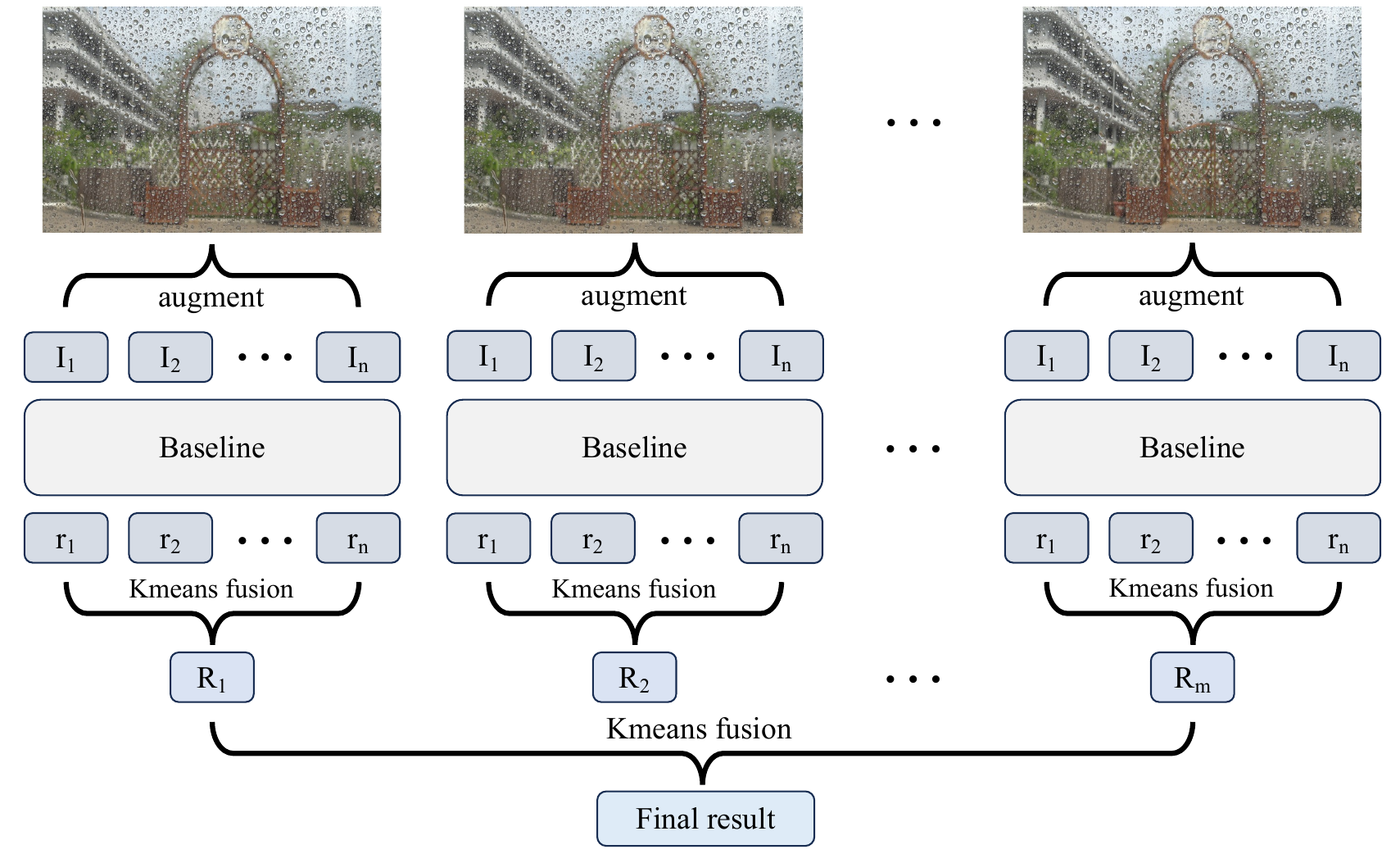}
    \caption{Dual kmeans fusion for RainDrop task \yeying{proposed by Team H3FC2Z}.}
    \label{fig:H3FC2Z}
\end{figure}
This team utilizes the RDDM~\cite{liu2024residualRDDM} as the deraining backbone and trained it with a patch size of $256\times 256$. Subsequently, the self-ensemble strategy used in the EDSR~\cite{lim2017enhancededsr} was improved by replacing the average ensemble with their ``Dual
Kmeans fusion" in the Fig.~\ref{fig:H3FC2Z}. Experimental results indicate that employing ``Dual Kmeans fusion"~\cite{chen2024ntirekmeans} increases the score of the RDDM baseline from 31.72 to 32.56. 

\noindent \textbf{Dual Kmeans Fusion.} 
Given a clean image $\mathcal{I}_{gt}$, several raindrop and blur degradations are added to $\mathcal{I}_{gt}$, obtaining $m$ images $(\mathcal{I}_{1}, \mathcal{I}_{2}, \cdots, \mathcal{I}_{m})$. As shown in Fig.~\ref{fig:H3FC2Z}, taking $\mathcal{I}_{1}$ as an example, they serve flipping and rotating as augmentations, generating $n$ images $(I_{1}, I_{2}, \cdots I_{n})$. \textbf{Stage-1}: The baseline model processes these images and obtains preliminary results $r_i = \text{Baseline}(I_i)$, where $i=1, 2, \cdots, n$. The images $r$ are flipped or rotated to a fixed angle and Kmeans fusion is performed to obtain $R_1$. \textbf{Stage-2}: After performing the above fusion on images $\mathcal{I}$, they get $m$ results $R_1, R_2, \cdots, R_m$. they perform Kmeans fusion on these $m$ results to get the final result image.

\noindent\textbf{Training and Testing Details.} The training dataset provided in this challenge is used for model training. To improve generalization, data augmentation techniques, including rotation and flipping are applied. The model is trained for 100,000 iterations using the AdamW optimizer with parameters $\beta_1 = 0.9$ and $\beta_2 = 0.95$, on a single NVIDIA A6000 GPU. Training is conducted with a batch size of 8, a learning rate of $3 \times 10^{-4}$, and a patch size of 256$\times$256. During inference, the authors adopt a Dual K-means fusion strategy to further enhance the model’s performance. Experimental results indicate that employing ``Dual Kmeans fusion" increases the score of RDDM~\cite{liu2024residualRDDM} baseline from 31.72 to 32.56. Specifically, Kmeans fusion in stage one increases the score of baseline from 31.72 to 32.26, and Kmeans fusion in stage two increases the score of baseline from 32.26 to 32.56.

\subsection{IIC Lab}

\begin{figure}[htbp]
    \centering
    \includegraphics[width=1.0\linewidth]{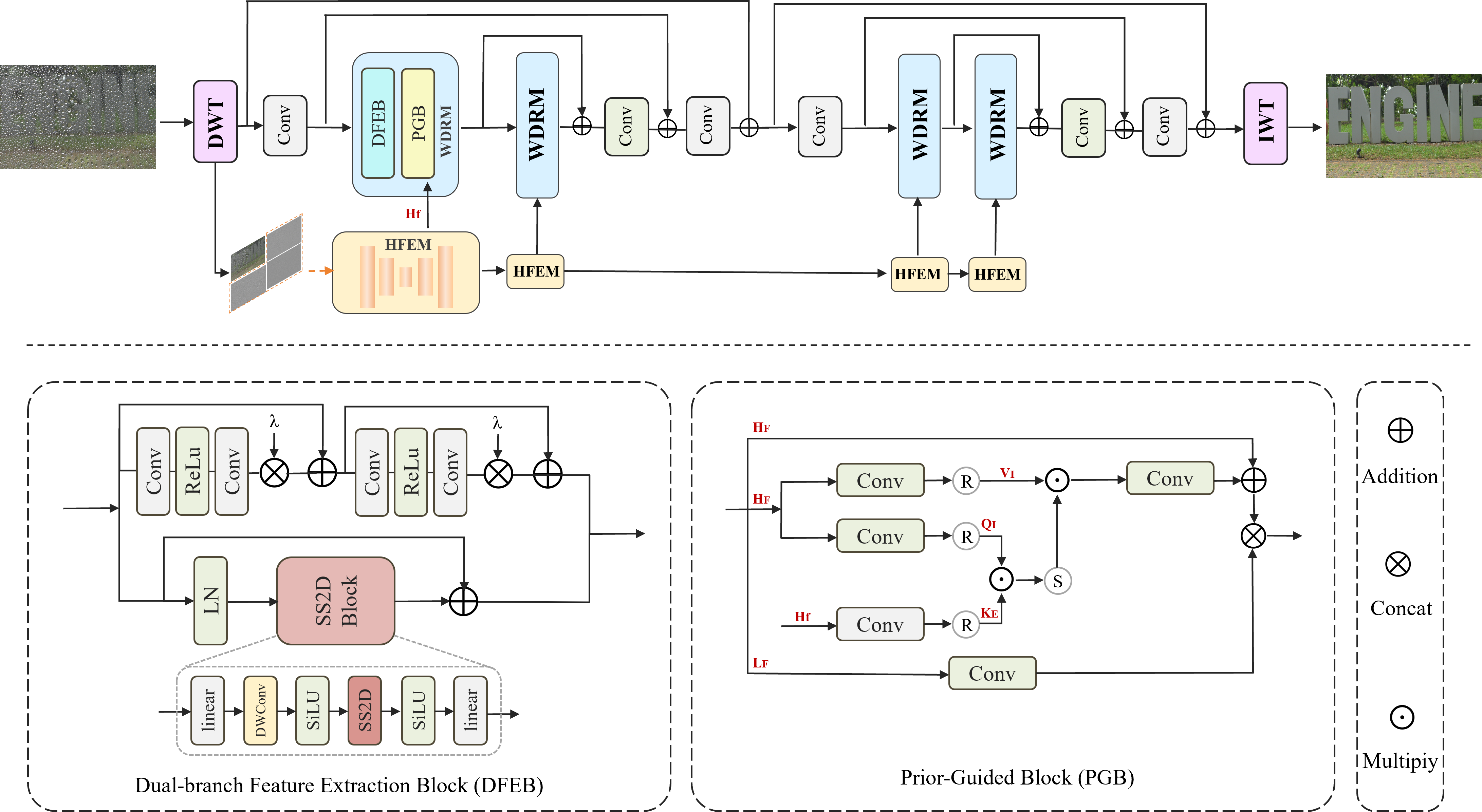}
    \caption{\yeying{The pipeline of the FA-Mamba proposed by Team IIC Lab}.}
    \label{fig:IICLab}
\end{figure}

This team developed an effective frequency-aware and Mamba-based network for image deraining, named FA-Mamba\yeying{, as shown in Fig.~\ref{fig:IICLab}}. Specifically, the key component of the proposed framework is the Wavelet Domain Restoration Module (WDRM) which contains a Dual-branch Feature Extraction Block (DFEB) that has superior local perception and global modeling capabilities and a Prior-Guided Module (PGM) that provides refined texture detail guidance for
feature extraction. It is worth mentioning that the refined texture details are obtained by enhancing the input high-frequency information through the High-Frequency Enhancement Module (HFEM).

\noindent\textbf{Training Details.}
During training, they utilized the Adam optimizer with a batch size of 1 and a patch size of 256 for a total of 80 iterations. The initial learning rate is fixed at $1e^{-4}$ for 60 iterations, and then decreased to $5e^{-5}$ for 20 iterations. No data augmentation techniques were applied. The entire framework is performed on PyTorch with an NVIDIA GeForce RTX 3090 GPU, which works in an end-to-end learning fashion without costly large-scale pertaining.

% BUPT CAT
\subsection{BUPT CAT}
\begin{figure}[htbp]
    \centering
    \includegraphics[width=1.0\linewidth]{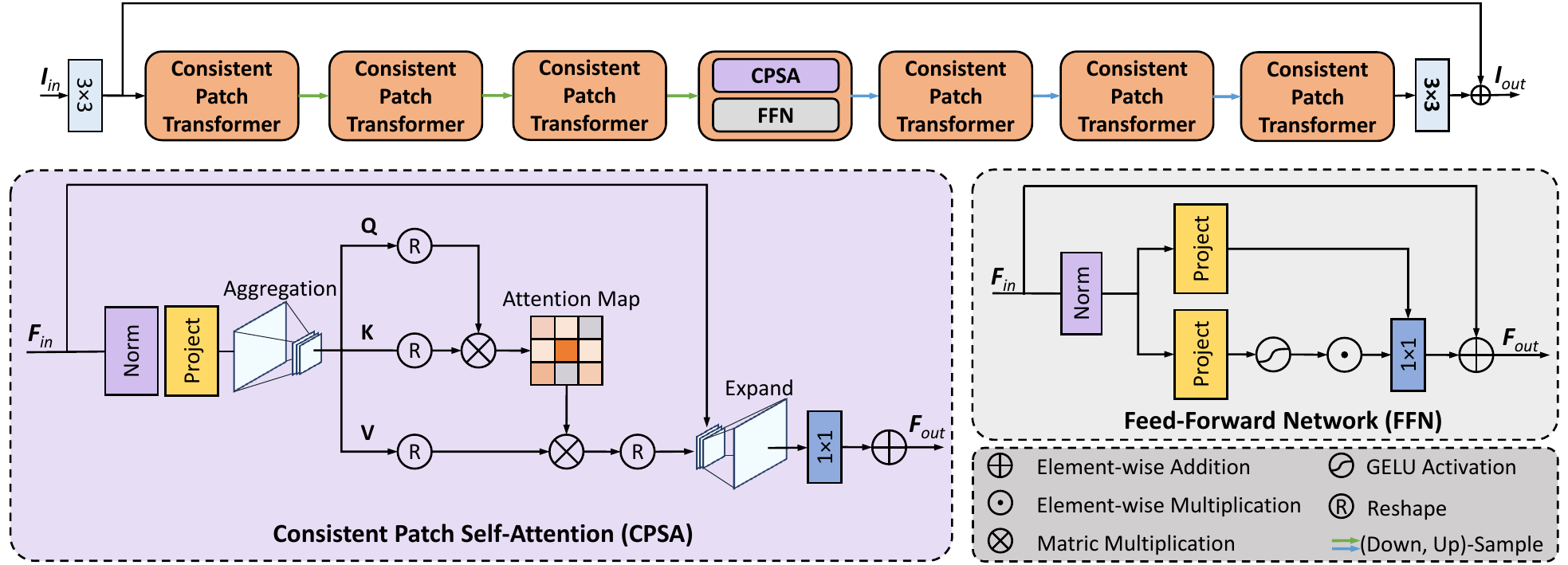}
    \caption{\yeying{The pipeline of the method proposed by Team BUPT CAT}.}
    \label{fig:BUPT_CAT}
\end{figure}
% \textcolor{red}{Please add citations in this sentence}

As shown in Fig.~\ref{fig:BUPT_CAT}, this team introduces a novel Consistent Patch Transformer (CPT) for dual-focused day and night raindrop removal task, which leverages a UNet-based architecture designed to enhance both spatial consistency and feature representation capability.
The framework comprises multiple Consistent Patch Transformer blocks, each consisting of two key components: Consistent Patch Self-Attention (CPSA) and a Feed-Forward Network (FFN). The model utilizes
the Test Time Local Converter (TLC) mechanism~\cite{chu2022improving} to effectively revisit global information aggregation and ensure robust and consistent feature learning with different patch sizes at training and testing.
The CPSA module is responsible for capturing both long-range dependencies and spatially consistent local details. Instead of using traditional window-based attention mechanisms, the CPSA module integrates a TLC-based feature aggregation and scaling strategy that maintains consistent patch sizes during training and testing, reducing spatial inconsistencies between training and testing. 

\noindent\textbf{Training Details.}
To reduce the training GPU memory, this team augments the input data by randomly cropping the input image into patches of the same size and performing strategies such as random rotation. 

\noindent\textbf{Testing Details.}
During the testing stage, in their self-attention part, they use the TLC strategy to segment and aggregate the full image into a series of patches of the same size as the training patch, and the rest of the model is the full image. This setup can effectively improve the inconsistency of the model’s patch size between training and testing, especially in the self-attention part.

\noindent\textbf{Implementation Details.}
This team utilizes the Pytorch framework with the NVIDIA GeForce RTX 4090. During training, they set the total batch size to 8, the initial learning rate from $5e^{-4}$ to $1e^{-5}$ with a scheduler in 500K iterations, and the patch size is set to 192$\times$192. For the loss function, they use $\mathcal{L}_1$loss and Fourier loss to constrain their model with weights of 1 and 0.1, respectively.
They train their framework using the Adam optimizer with $\beta1=0.9, \beta2=0.99$.
They set the number of channels to 64 in their network. The total training duration is approximately 80 hours. The training and test sets are official datasets provided by the dual-focused day and night raindrop removal challenge.

\subsection{WIRTeam}
\begin{figure}[htbp]
    \centering
    \includegraphics[width=0.8\linewidth]{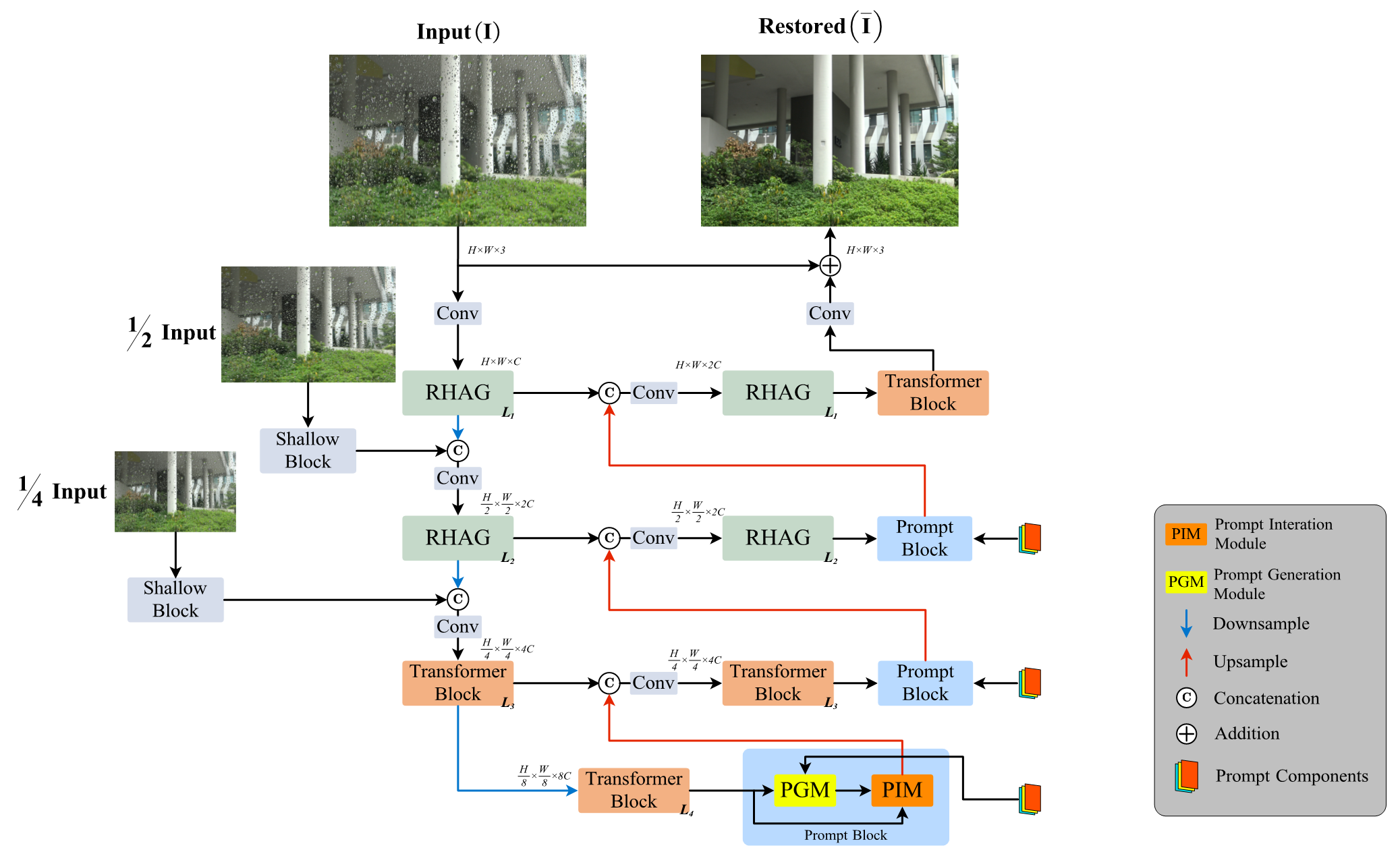}
    \caption{\yeying{The pipeline of the method proposed by Team WIRteam}.}
    \label{fig:WIRTeam}
\end{figure}
Inspired by recent advancements~\cite{li2024promptcir,chen2023activating,potlapalli2023promptir,zamir2022restormer} in image restoration and image deraining, \yeying{this team} propose a novel multi-scale prompt-based image deraining approach (MPID) that incorporates both local and global attention mechanisms\yeying{, as shown in Fig.~\ref{fig:WIRTeam}.}
Specifically, given a degraded image $I \in R^{H \times W \times 3}$
, their model produces a corresponding clear image $\bar{I} \in R^ {H\times W \times 3}$ through a 4-level encoder-decoder framework. In the shadow two layers, they employ Residual Hybrid Attention Groups (RHAG)~\cite{chen2023activating} to capture detailed local features. For deeper layers, they integrate Transformer Blocks~\cite{zamir2022restormer} to facilitate cross-channel global feature extraction.
In the encoding phase, acknowledging the advantages of utilizing images at various resolutions for deraining tasks~\cite{chen2024bidirectional}, they additionally incorporate multi-scale image information (at 1/2 and 1/4 of the original resolution) to enhance auxiliary information during the encoding process. During decoding, considering the dual focus on raindrop-focus and background-focus images within the Raindrop Clarity dataset~\cite{ntire2025day}—which introduces both raindrop occlusions and background blurring—they introduce specialized
prompt mechanism~\cite{li2024promptcir}. These components are designed to address and decouple different types of degradation factors. Notably, Prompt Block, which integrates Prompt Generation Module (PGM) and Prompt Interaction Module (PIM), utilizes image-specific cues to effectively guide the image reconstruction process, thereby improving the clarity and quality of the restored images.\\
\noindent\textbf{Training Details.}
They employ the end-to-end training methodology, training their model for 400 epochs. The training procedure is based on the AdamW optimizer with the decay parameters $\beta_1 = 0.9$ and $\beta_2 = 0.99$. The initial learning rate is 2e-4 and gradually reduces to 1e-6 with the cosine annealing strategy. Horizontal and vertical flips are adopted for data augmentation. Furthermore, their training is merely based on the Raindrop Clarity dataset provided by the competition organizers, without the use of any extra datasets or pretrained models.\\
\noindent\textbf{Testing Details.}
During the testing phase, they preprocess the input images by padding them to the multiple of 32. This ensures that the dimensions of the input images are compatible with the architecture of their method. They don’t resort to any other means of Test-Time Augmentation during the testing phase.\\

% GURain
\subsection{GURain}
\begin{figure}[htbp]
    \centering
    \includegraphics[width=1.0\linewidth]{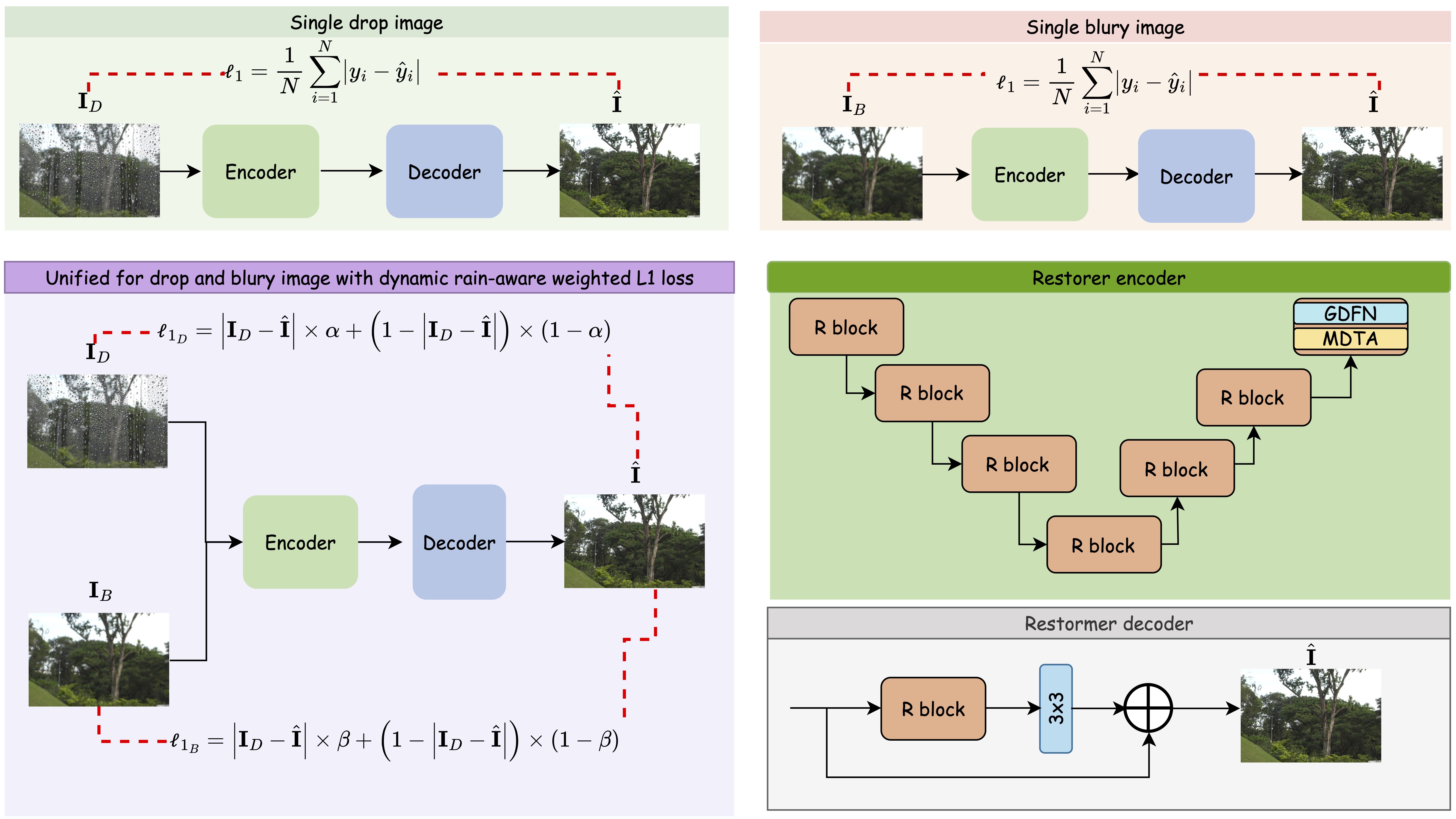}
    \caption{\yeying{Comparison between the single-degradation Restormer and the dual-degradation method proposed by Team GURain. The top row shows the original Restormer handling drop- or blur-degraded images separately using standard L1 loss. The method by Team GURain (bottom left) unifies both degradations within a single encoder-decoder framework and introduces a dynamic, rain-aware weighted L1 loss to better emphasize challenging regions. The architecture (bottom right) retains the Restormer backbone while enhancing its ability to handle both degradations jointly.}}
    \label{fig:GURain}
\end{figure}
\begin{figure}[htbp]
    \centering
    \includegraphics[width=1.0\linewidth]{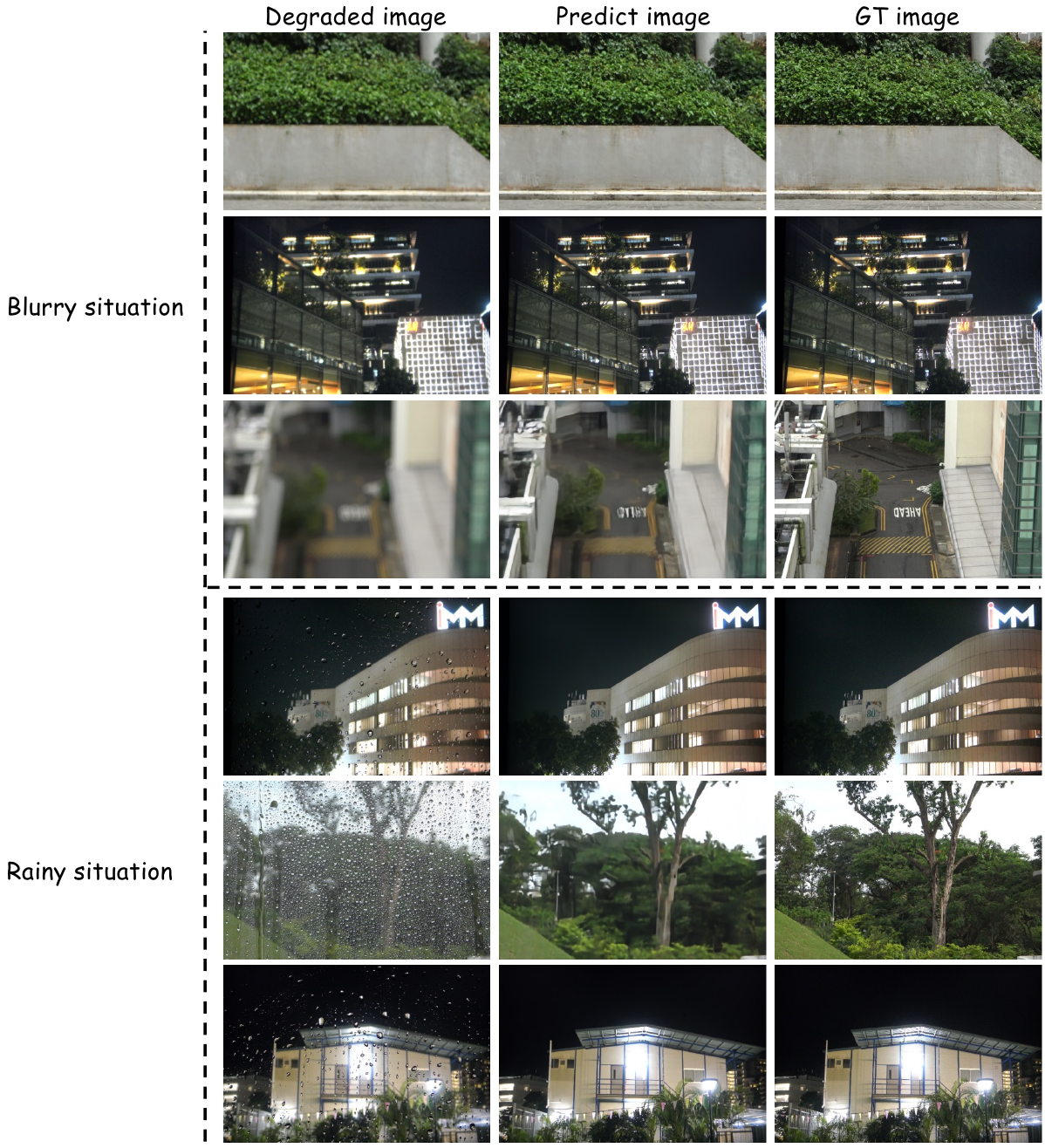}
    \caption{\yeying{Visualization of the method proposed by GURain on validation image. The top three rows are about deblurry. The bottom three rows are about deraining. Columns from left to right are degraded image, predicted image and ground truth image.}}
    \label{fig:GURain_result}
\end{figure}

This team addresses the dual degradation challenge—rain and blur—by unifying the training data into a single framework using the vanilla Restormer, originally designed for single degradation restoration, and by introducing a dynamic rain-aware weighted L1 loss\yeying{, as shown in Fig.~\ref{fig:GURain}}. 
Instead of employing a dual-input network that struggles to differentiate between blur and rain, they merge rainy and blurry images into one input paired with a clear ground truth, allowing Restormer to learn a common mapping for both degradations. Moreover, recognizing that rain streaks affect only parts of an image, their adaptive loss function assigns higher weights to rainy regions and lower weights to clean areas, thereby guiding the network to focus on the more challenging parts of the image. This integrated approach leads to effective deraining while simultaneously mitigating blur, resulting in improved overall image restoration performance as shown in Fig.~\ref{fig:GURain_result}.

\noindent\textbf{Training and Testing Details.} They train the default Restormer~\cite{zamir2022restormer} using a custom dynamic, rain-aware weighted $\mathcal{L}_1$ loss on merged rainy/blurry inputs paired
with clear ground truths, with progressive patch size scaling. The
custom loss emphasizes challenging rainy regions, yielding improved quantitative metrics and visually cleaner, less blurred results while preserving
the efficiency of Restormer~\cite{zamir2022restormer}. During inference, the standard Restormer pipeline processes single degraded images.

\subsection{BIT\_ssvgg}
\begin{figure}[htbp]
    \centering
\includegraphics[width=1.0\linewidth]{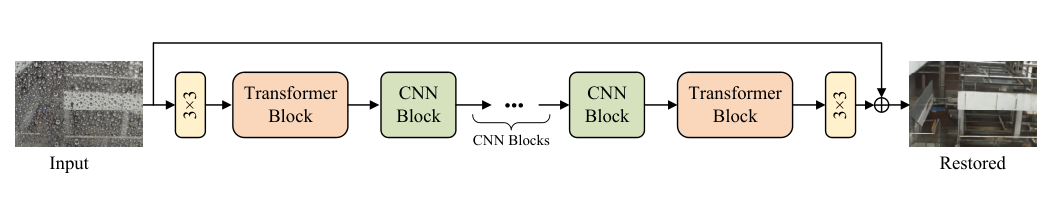}
    \caption{\yeying{The pipeline of the method proposed by Team BIT\_ssvgg}.}
    \label{fig:BIT_ssvgg}
\end{figure}

\begin{figure}[htbp]
    \centering
    \includegraphics[width=1.0\linewidth]{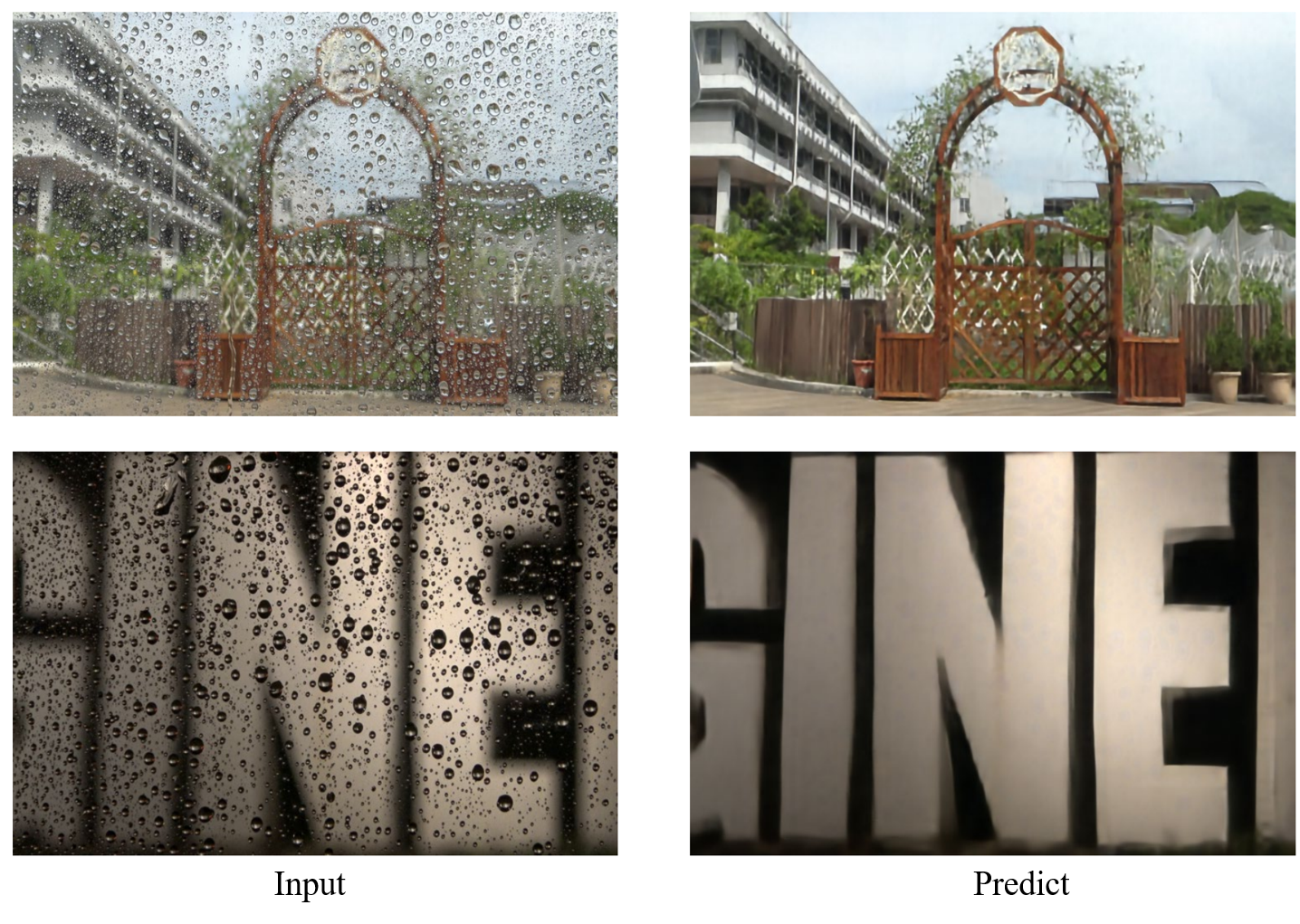}
    \caption{\yeying{Visualizations of partial image restoration results by Team BIT\_ssvgg.}}
    \label{fig:BIT_ssvgg_result}
\end{figure}

As illustrated in Fig~\ref{fig:BIT_ssvgg}, their method is designed to operate on rain-degraded images and generate the corresponding clean outputs. It follows an encoder–decoder architecture. Given an input degraded image, they first apply a 3$\times$3 convolution to extract initial features, which are then processed by a Transformer module with a global receptive field. Since the rain streaks often appear in large and unevenly distributed regions, it is essential for the network to capture long-range dependencies. To address this, they adopt the attention mechanism from Restormer~\cite{zamir2022restormer}, which enables efficient global context modeling with reduced computational complexity. Subsequently, the features are fed into a series of CNN-based modules to enhance local feature representation and compensate for the Transformer’s limited ability to model fine-grained structures. Their CNN blocks are built upon NAFNet~\cite{chen2022simpleNAFNet}, chosen for its lightweight design and effectiveness. After that, the features are further refined through another Transformer block and finally passed through a 3$\times$3 convolution layer. The output is then added to the input image in a residual manner to produce the restored image. To preserve spatial information during encoding and decoding, they employ pixel-unshuffle and pixel-shuffle operations for downsampling and upsampling, respectively, after each CNN block. This prevents information loss during resolution changes. By integrating both Transformer and CNN components, their hybrid architecture leverages the strengths of each: global context aggregation and local detail preservation. Moreover, due to its improved generalization ability and reduced tendency to overfit on small datasets, the model can be effectively trained solely on the provided benchmark dataset without requiring any additional data.

\noindent\textbf{Implementation details.} Their model is implemented in Python using the PyTorch framework (version 1.13.1). The training is conducted on NVIDIA RTX 4090. They adopt the AdamW optimizer with $\beta_1 = 0.9$ and $\beta_2 = 0.9$. The learning rate is scheduled using a cosine annealing strategy, gradually decaying from $1\times e^{-3}$ to $1\times e^{-7}$ over the course of training. The training dataset is exclusively provided by the competition organizer, and no additional external data is used. They trained the model for 300 epochs (approximately 30 hours) with a batch size of 8 and a patch size of 256$\times$256. Standard data augmentation techniques are applied, including random horizontal/vertical flipping and random rotations to improve generalization.

\subsection{CisdiInfo-MFDehazNet}

\begin{figure}[htbp]
    \centering
\includegraphics[width=1.0\linewidth]{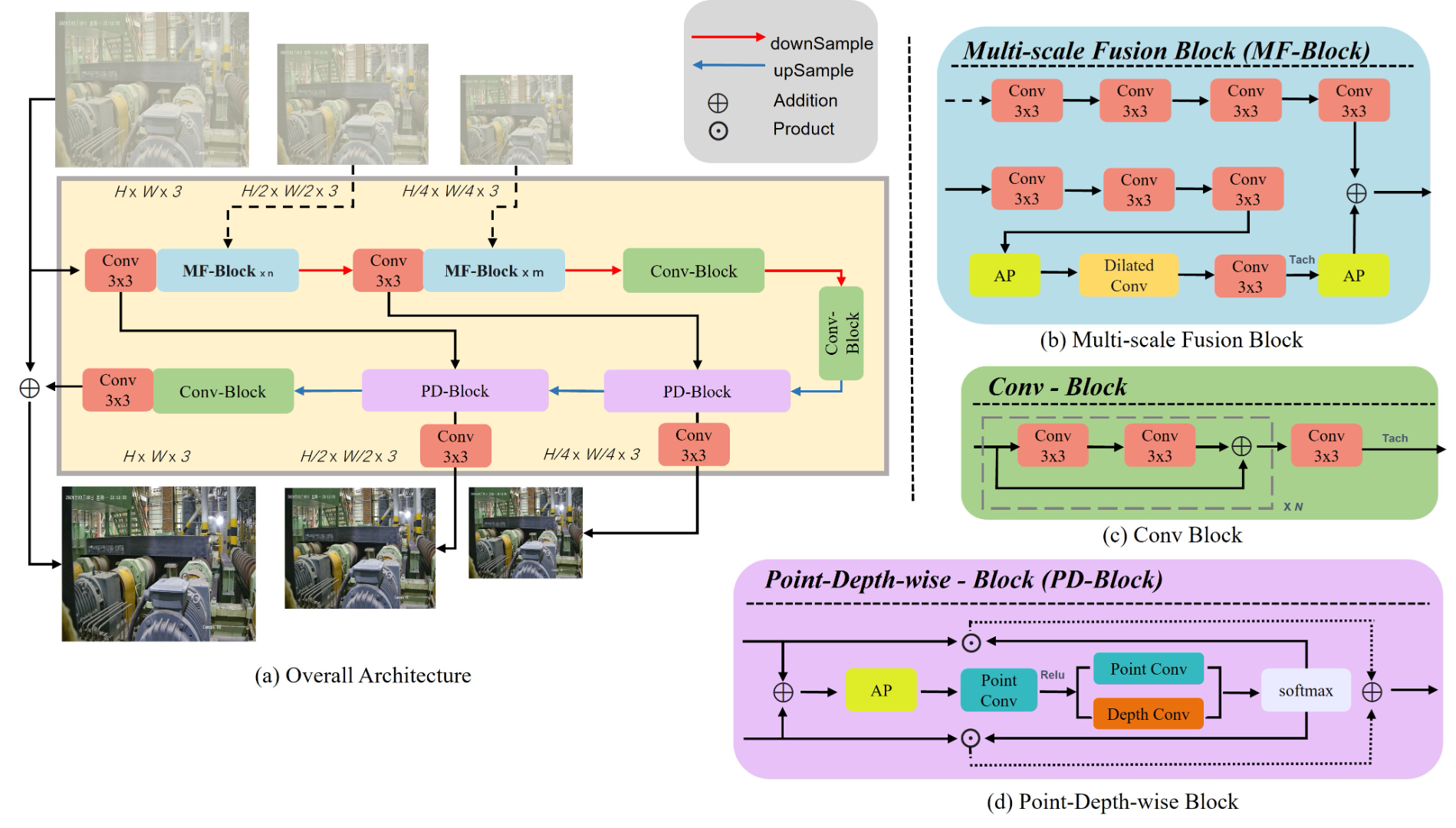}
    \caption{\yeying{Framework of the method proposed by Team CisdiInfo-MFDehazNet}.}
    \label{fig:CisdiInfo-MFDehazNet}
\end{figure}
\yeying{In Fig.~\ref{fig:CisdiInfo-MFDehazNet}, t}his team proposes an effective and lightweight model that can be applied in the industrial site, named MFDehaz-Net, which has two specially designed components, \ie, Multi-scale Fusion Block and Point-Depth wise Block, helping it achieve deep fusion of image features of different scales to obtain better global understanding following~\cite{song2023visionCisdiInfo-MFDehazNet,gao2024efficientCisdiInfo-MFDehazNet}.

\textbf{Training description}
The model proposed is implemented with Pytorch1.10.2+CUDA11.3 and trained for 1000 epochs on an NVIDIA GeForce RTX3090 GPU. The batch
size and learning rate are set as 16 and 0.0008,respectively, the number
of warmup epochs is 50, and Adam is selected as the optimizer. Random
rotation and horizontal inversion are also used as data augmentation
methods.\\
\textbf{Testing description}
During testing, all the images are resized to 720x480 and then fed into the
loaded pretrained model. 
MFDehaz-Net has two specially designed components, i.e., Multi-scale Fusion Block (MF-Block) and Point-Depth wise Block (PD-Block), which help MFDehaz-Net achieve deep fusion of image features of different scales to obtain better global features. Besides, to reduce the damage to the original color of the image caused by image dehazing, MFDehaz-Net integrates the supervision signal in the frequency domain with a specially designed loss function.\\

\subsection{McMaster-CV}
\begin{figure}[htbp]
    \centering
    \includegraphics[width=1.0\linewidth]{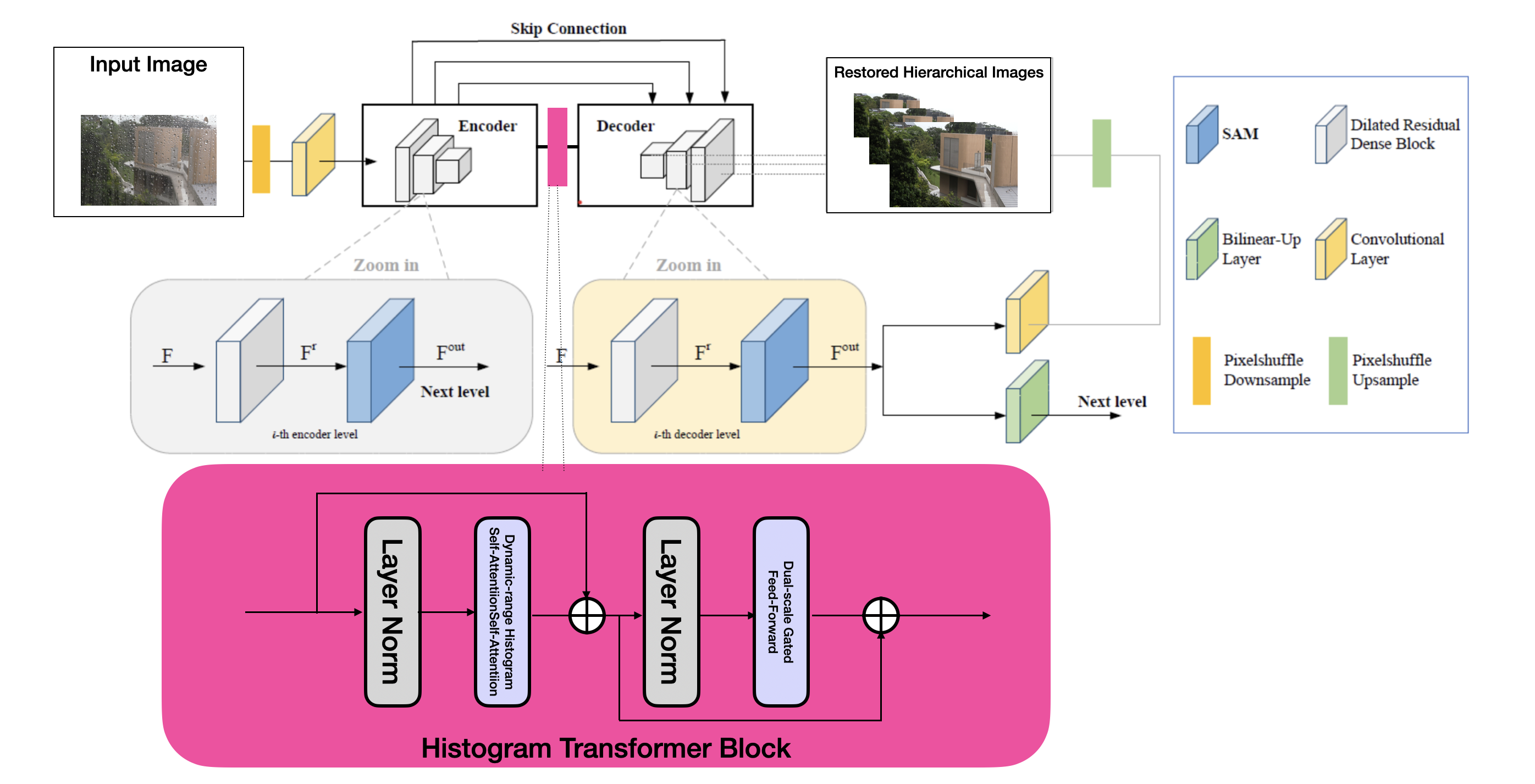}
    \caption{\yeying{Framework of the method proposed by Team McMaster-CV}.}
    \label{fig:McMaster-CV}
\end{figure}
As shown in the Fig.~\ref{fig:McMaster-CV}, the framework of this team is based on ESDNet \cite{yu2022towards}. The backbone primarily consists of an encoder-decoder network. At each encoder and decoder level, a Semantic-Aligned Scale-Aware Module (SAM) is incorporated to address scale variations. Additionally, this team introduces a Histogram Transformer Block \cite{sun2024restoring}, which employs histogram self-attention with dynamic range spatial attention. This block is placed between the encoder and decoder to achieve global and efficient degradation removal.

Besides, for network training, they designed their loss function based solely on a single level, $\hat{I}_{1}$, which corresponds to the original image resolution:
\begin{equation}
 {L}_{loss}={L}_{C}\left(I, \hat{I}\right) + {\lambda}{L}_{Percep}\left(I, \hat{I}\right)
\label{eq1} 
\end{equation}
where ${L}_{1}$ and ${L}_{Percep}$ represent Charbonnier loss~\cite{lai2018fast}, and perceptual loss~\cite{johnson2016perceptual}, respectively. The weighting factor is set as $\lambda = 0.04$.

\noindent\textbf{Training and Testing Details.}
This team trained their model on a single NVIDIA 1080Ti(12GB VRAM) GPU. During training, They set the batch size to 3 and the patch size to 480. For the training strategy, following Restormer\cite{zamir2022restormer}, they adopted the \textit{CosineAnnealingRestartCyclicLR} scheduler, which adjusts the learning rate using a cosine annealing schedule with restarts to promote better convergence and escape local minima. Specifically, the training consists of two cycles with periods of 46{,}000 and 104{,}000 iterations, respectively, with minimum learning rates of 0.0003 and 0.000001. The learning rate is reset to its initial value at the beginning of each cycle. To enhance generalization, they incorporated \textit{mixup}-based data augmentation, where training samples are linearly combined using a Beta distribution with a shape parameter of 1.2. Additionally, they enabled the use of identity mapping to retain some original samples during training. The generator is optimized using the \textit{Adam} optimizer with a learning rate of $2 \times 10^{-4}$ and standard momentum parameters ($\beta_1 = 0.9$, $\beta_2 = 0.999$).
 During the testing phase, they directly fed the input into their model to obtain the final output.
 
\subsection{Falconi}
\begin{figure}[htbp]
    \centering
    \includegraphics[width=1.0\linewidth]{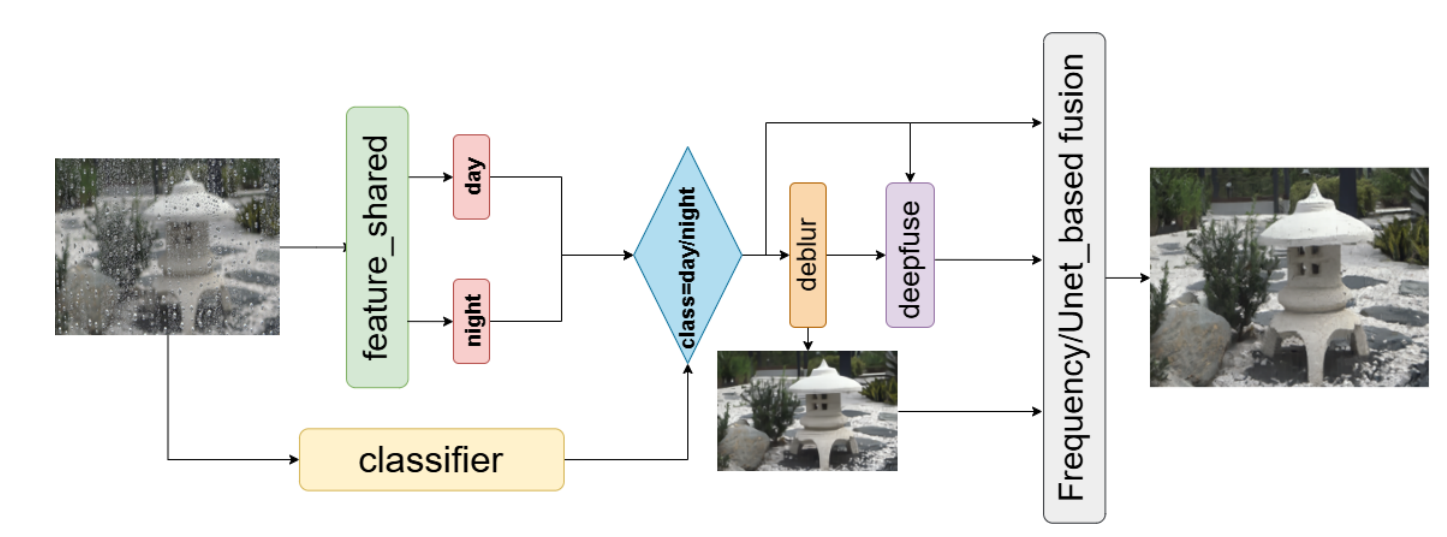}
    \caption{The proposed unified framework for joint image deraining and deblurring by Team Faconi. The framework is composed of four stages: Context Classification, Adaptive Deraining, Multi-Task Feature Extraction, and Iterative Deblurring with Adaptive Fusion.}
    \label{fig:Falconi}
\end{figure}

This team presents a unified framework for image deraining that synergistically combines classification, multi-task learning, and adaptive fusion. Their approach leverages pre-trained models to manage complexity and enhance performance. Specifically, a MobileNetV2-based classifier initially categorizes images as day or night, guiding subsequent processing. A pre-trained Diffusion Transformer(DiT), fine-tuned for both day and night scenarios, serves as a core component for night image adaptation. For deblurring, they incorporate FFTformer. Finally, they explore both traditional and U-Net-based fusion strategies to combine intermediate outputs. The entire framework is trained using a combination of curated external datasets for day/night classification and DiT adaptation, alongside dedicated datasets for deblurring and synthetically generated samples from intermediate deraining stages.

As shown in Fig.~\ref{fig:Falconi}, their training pipeline comprises four interconnected stages:

\noindent\textbf{Stage 1. Day/Night Classification.}
They initialize a MobileNetV2-based classifier to discriminate between day and night scenes, providing a foundational context for subsequent stages.

\noindent\textbf{Stage 2. Adaptive DiT Model.}
They extend a pre-trained DiT~\cite{peebles2023scalableDiT} model, initially designed for night imagery, by fine-tuning it on a mixed dataset of day and night images. This stage incorporates the pre-trained classifier (from Stage 1) with a dynamic weighting strategy, enabling context-aware learning based on the time of day.

\noindent\textbf{Stage 3. Multi-Task Branching Network.}
They construct a three-branch multi-task network. The MobileNetV2 classifier (from Stage 1) forms one branch. The penultimate layer of the adapted DiT model (from Stage 2) serves as a shared feature backbone. Two parallel, final-layer branches, specialized for day and night conditions respectively, are instantiated. During training, the shared backbone and the classification network remain frozen; optimization is restricted to the day- and night-specific branches. The classifier output directly dictates the active branch for a given input.

\noindent\textbf{Stage 4. Iterative Deblurring Refinement.}
They integrate the FFTformer deblurring network~\cite{kong2023efficientFFTdebluring} using a two-round fine-tuning process. The first round utilizes a dataset of blurred images. The second round leverages a synthetic dataset generated from the deraining outputs of the DiT model (from Stage 2), thereby aligning the deblurring process with the specific characteristics of the derained images.

During inference, the input image undergoes a two-stage process: deraining followed by deblurring. First, the image is processed by the deraining pipeline (Stage 3), and subsequently enhanced by the deblurring network (Stage 4). To effectively integrate the complementary information from these stages, they employ a dual-fusion approach:

\noindent\textbf{Frequency-Domain Fusion.}
This method combines derained and deblurred images by adaptively weighting their frequency components based on sharpness, noise, and edge information, prioritizing low-frequency content from the derained image and high-frequency details from the deblurred image.

\noindent\textbf{Learned Fusion.}
This method utilizes a U-Net-based neural network~\cite{liu2022robustU-Net}, trained on a diverse dataset of model outputs, to learn a non-linear mapping that optimally fuses the derained and deblurred images, implicitly addressing the weaknesses of each individual processing stage. The table \ref{tab:Falconi} presents the final evaluation results of their image deraining and deblurring framework, with validation and test values reported. The validation score of 33.05 indicates the model’s strong generalization capability during the training phase, while the test score of 31.71 reflects its performance on previously unseen data. The relatively small difference between the validation and test results suggests that the model has effectively avoided overfitting, main-taining robust performance across different datasets. These results validate the effectiveness of the unified framework, which integrates day/night classification, adaptive DiT model, multi-task branching, and iterative deblurring refinement.

This team developed their method based on the Dit model, utilizing the Adam optimizer with a learning rate of $1\times e^{-5}$, and executed the training process on a single NVIDIA 4090D GPU (24GB). The proposed Raindrop Clarity dataset was exclusively employed, with pairs of (drop, clear) used in stages 1 and 2, and pairs of (blur, clear) utilized in stage 3. In Stage 1, they trained a binary classifier, achieving convergence in under 30 minutes. Stage 2 required 25 hours of training, while Stage 3 took 14 hours. Finally, Stage 4 involved training a U-Net, which was completed in less than 10 minutes. Throughout the training process, learning rate optimization was employed as part of the training strategy.

\begin{table}
\centering
\begin{tabular}{cc}
\hline
Dataset & Value \\
\hline
Validation & 33.05291 \\
Test & 31.70666 \\
\hline
\end{tabular}
\caption{Final evaluation results of the Image Deraining and Deblurring Framework by Team Falconi.}
\label{tab:Falconi}
\end{table}

\subsection{Dfusion}

\begin{figure}[htbp]
    \centering
    \includegraphics[width=1\linewidth]{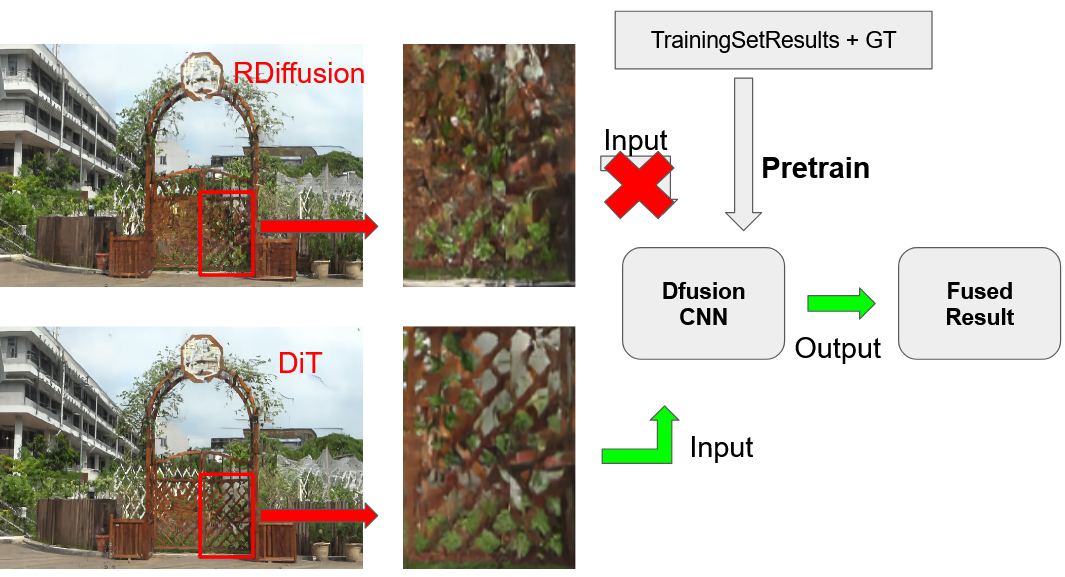}
    \caption{The implementation structure of Team Dfusion.}
    \label{fig:Dfusion}
\end{figure}

The proposed \textbf{Dfusion} method is a dual-branch fusion framework that leverages the complementary strengths of two state-of-the-art pretrained models—RDiffusion~\cite{ozdenizci2023restoringrdiffusion} and DiT~\cite{peebles2023scalableDiT}—to effectively remove raindrops from dual-focused images. As illustrated in Fig.~\ref{fig:Dfusion}, the method first generates two intermediate restored images, each excelling in different visual aspects (for example, one may deliver smooth textures while the other preserves fine details).

These intermediate outputs are then stacked and passed through a lightweight fusion CNN, which is specifically trained to combine the best qualities of both inputs. The final model is optimized using a joint loss function that integrates PSNR, SSIM, and LPIPS, ensuring both high quantitative accuracy and strong perceptual quality.

To enhance computational efficiency, input images are processed in overlapping patches. These patches are restored separately by the pretrained models and then reassembled prior to the fusion step. This patch-based strategy not only speeds up inference but also helps preserve local details across the image.

\begin{figure}[htbp]
    \centering
    \includegraphics[width=1\linewidth]{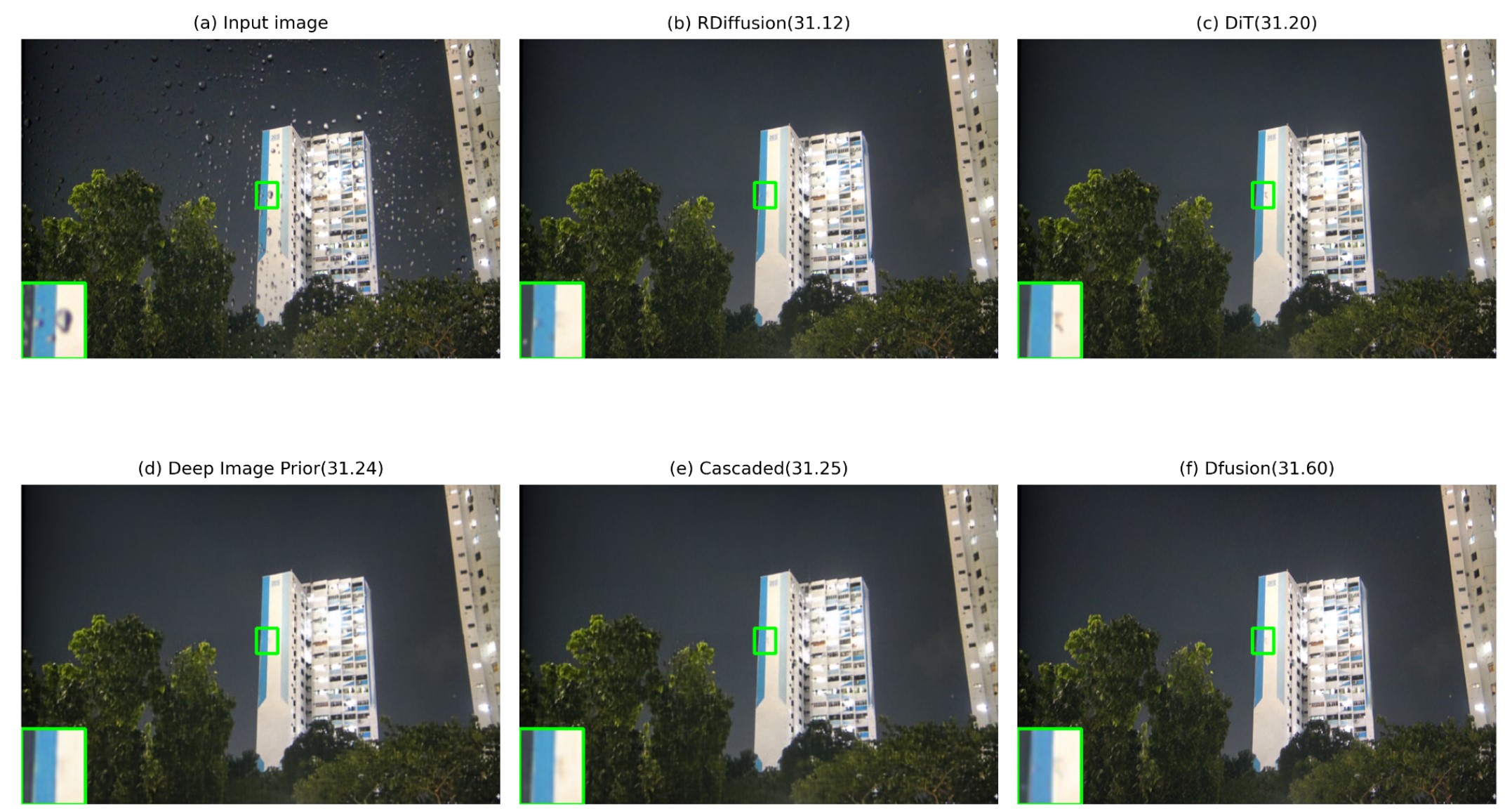}
    \caption{The result comparison of Team Dfusion.}
    \label{fig:Dfusion2}
\end{figure}

Their fusion approach draws inspiration from mixture-of-experts strategies \cite{frick2025prompttoleaderboard}, adapting them to the raindrop removal task. By fusing complementary features from two high-performance models, Dfusion is able to achieve robust restoration results in both daytime and nighttime scenarios, as shown in Fig. \ref{fig:Dfusion2}.

\noindent\textbf{Training Details.}
The fusion network is trained using a single RTX4060 8GB laptop. Adam optimizer is used with an initial learning rate of $1\times10^{-4}$. During training, each input image (of resolution 720$\times$480) is split into six overlapping patches of size 256$\times$256. The model is then optimized with a joint loss function that combines PSNR, SSIM, and LPIPS.

\noindent\textbf{Testing Details.}
In the inference stage, intermediate outputs from RDiffusion and DiT are first generated. These outputs are concatenated into a multi-channel tensor and subsequently processed by the lightweight fusion CNN. This process results in a final restored image that effectively preserves both global structure and fine local details.

\subsection{RainMamba}
% Done!
\begin{figure}[htbp]
    \centering
    \includegraphics[width=1.0\linewidth]{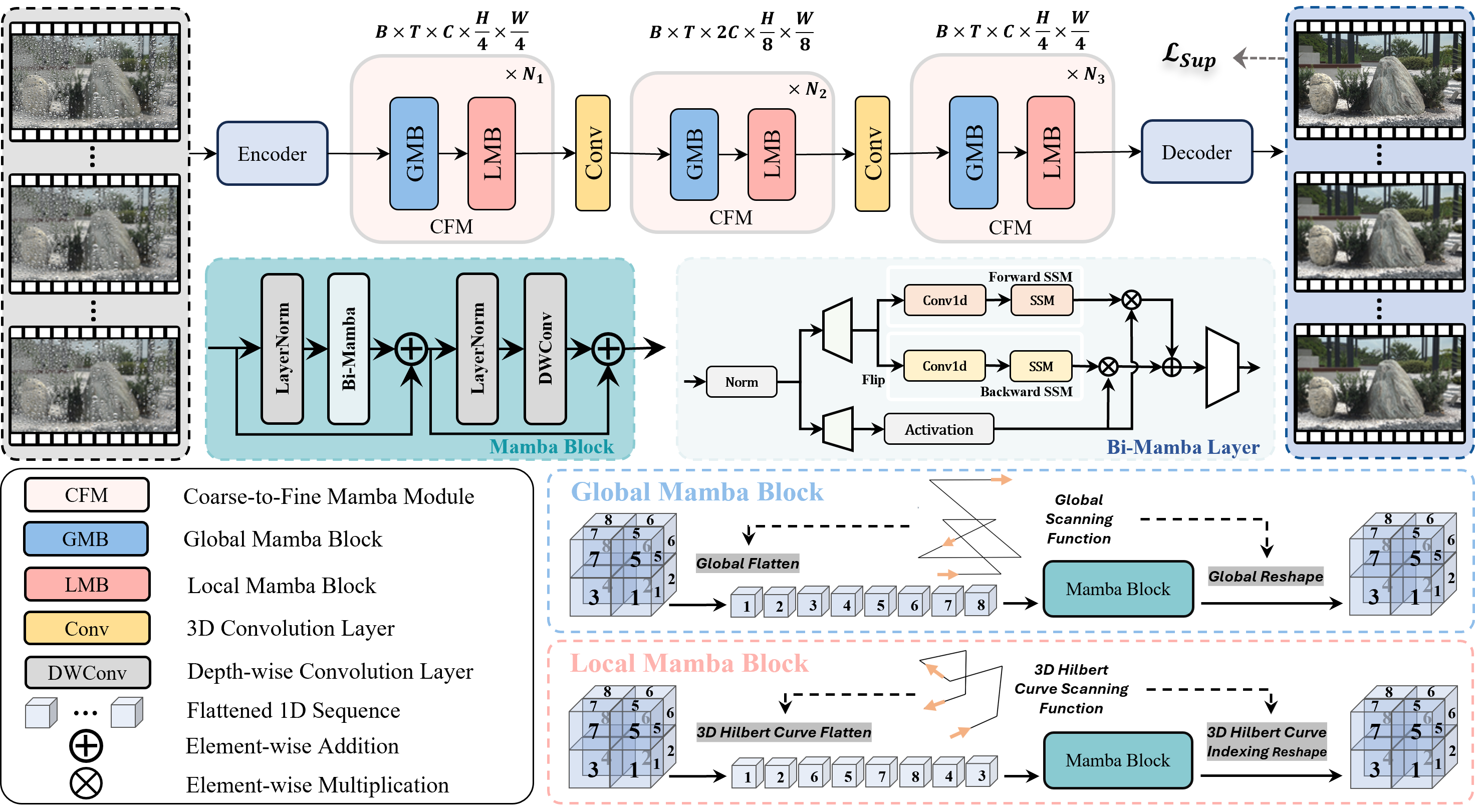}
    \caption{Network architecture of Team RainMamba.}
    \label{fig:RainMamba}
\end{figure}

The team proposes a video restoration framework~\cite{wu2024rainmamba} to adapt state space models to Day and Night Raindrop Removal tasks\yeying{, as shown in Fig.~\ref{fig:RainMamba}}. Their approach applies a global scanning mechanism to causally process the temporal data with linear complexity. The core innovation of their method lies in equipping State Space Models (SSMs) with a Hilbert scanning mechanism, which achieves localized scanning across both temporal and spatial dimensions. Given a sequence of rainy video frames, their cascading Coarse-to-Fine Mamba Module (CFM) receives the encoded features as input and causally models temporal corrections using improved state space models. The CFM employs Global Mamba Block (GMB) and Local Mamba Block (LMB) to capture sequence-level global and local spatio-temporal dependencies. They develop a novel Hilbert scanning paradigm in LMB to promote the Mamba's locality learning. To enhance the visual quality of the restored results, they adopt a combination of PSNR loss and perceptual loss in their training process.

\noindent\textbf{Training Details.}
Their network is trained on NVIDIA RTX 4090 GPUs and implemented on the Pytorch platform. At each training iteration, the input frame is randomly cropped to a spatial resolution of 256$\times$256, and the number of frames per video clip is 2. The total number of training iterations is 300k. They adopt the Adam optimizer and the polynomial scheduler with a power of 1.0. The initial learning rate of their network is set to $5\times 10^{-4}$ with a batch size of 8 and a warm-up start of 2k iterations. 

\noindent\textbf{Testing Details.}
In the testing phase, they take two frames of each video as a segment as input and input the full size of each frame.

\subsection{RainDropX}
% Done!
\begin{figure}[htbp] 
\centering 
\includegraphics[width=1.0\linewidth]{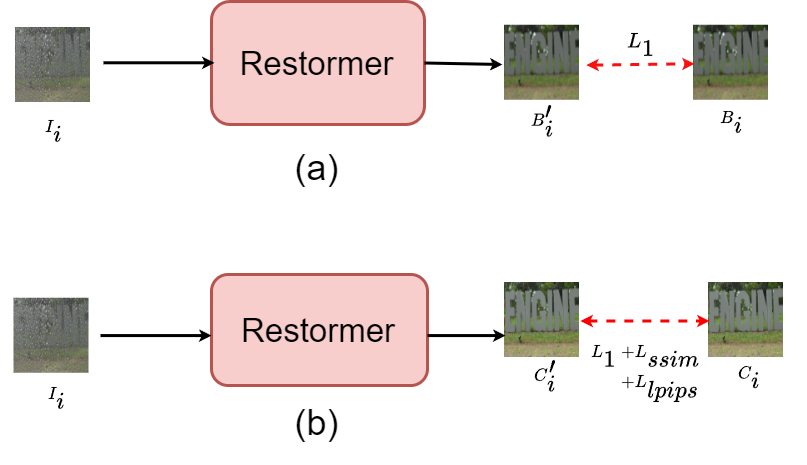} 
\caption{The diagram of the proposed method by Team RainDropX.} 
\label{fig:RainDropX} 
\end{figure}

The team proposes an enhanced version of Restormer \cite{zamir2022restormer}, introducing additional constraints for improving the perceptual and structural integrity of the restored images. Unlike the original Restormer model, which was trained for individual restoration tasks using $L_1$ loss, the proposed method incorporates LPIPS \cite{zhang2018lpips} and SSIM \cite{wang2004imagessimloss} losses to preserve perceptual quality, edge preservation, and texture consistency. The training process is conducted in two stages, starting with fine-tuning the pretrained weights for blurred output prediction and then transitioning to predicting clean images with added perceptual losses. The overall framework is shown in Fig.~\ref{fig:RainDropX}.

\noindent\textbf{Training Details.} The model is trained using the pretrained Restormer weights with a two-stage training approach. In Stage 1, $L_1$ loss is used to predict blurred outputs from degraded inputs containing both raindrops and blur. In Stage 2, LPIPS and SSIM losses are introduced alongside $L_1$ loss to optimize clean image restoration. The training is conducted on the Raindrop Clarity dataset~\cite{jin2024raindrop}, and the model is trained using a single GPU.

\subsection{Cidaut AI}
% Done!
\begin{figure}[htbp]
    \centering
    \includegraphics[width=1.0\linewidth]{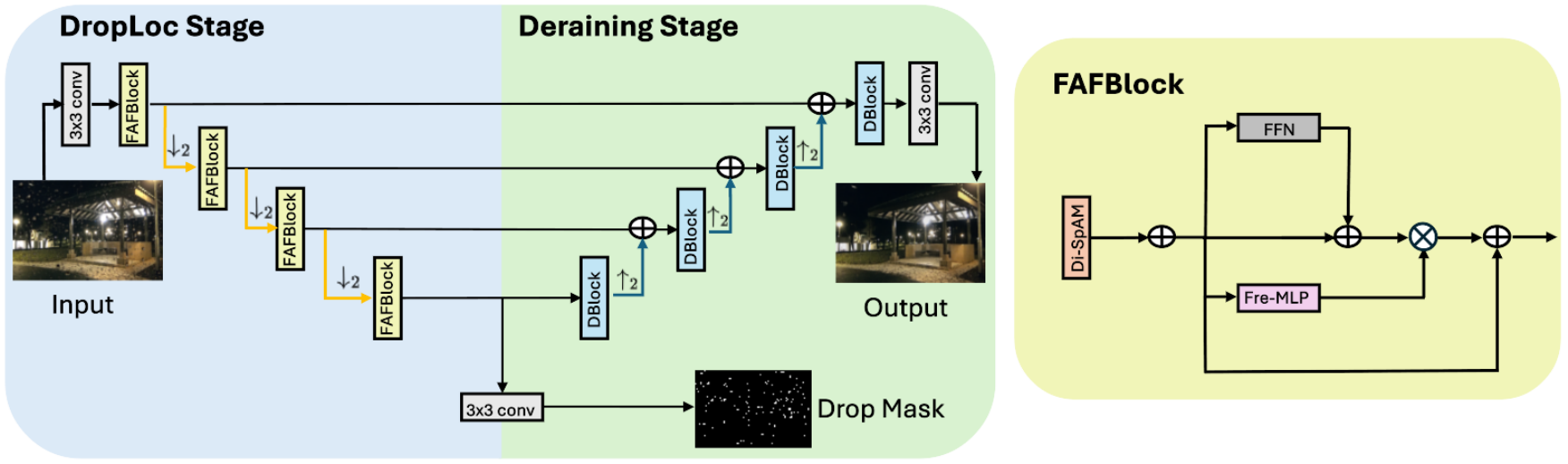}
    \caption{The proposed DropFIR by Team Cidaut AI.}
    \label{fig:Cidaut_AI}
\end{figure}

This team proposes DropFIR, an encoder-decoder architecture inspired by DarkIR~\cite{feijoo2024darkir}, tailored for raindrop removal\yeying{, as shown in Fig.~\ref{fig:Cidaut_AI}.} The model introduces a DropLoc stage based on a custom Fourier-Attention-Fusion Block (FAFBlock), adapted from FLOL~\cite{benito2025flol}, to extract a spatial drop mask. The predicted mask guides the Deraining Stage, enabling the decoder to remove localized drops and blur for image restoration.

\noindent\textbf{Training Details.} The model is trained using the AdamW optimizer ($\beta_1=0.9$, $\beta_2=0.999$, weight decay 0.01) with a cosine annealing learning rate schedule starting at $10^{-3}$ and decaying to $10^{-7}$. Training is conducted in two phases: a 100-epoch pretraining on $320\times480$ random crops, followed by 300 epochs on full-resolution images. The model is trained on four NVIDIA H100 GPUs with a batch size of 16, using only the Raindrop Clarity dataset~\cite{jin2024raindrop}.

\subsection{DGL\_DeRainDrop}
\begin{figure}[htbp]
    \centering
    \includegraphics[width=1.0\linewidth]{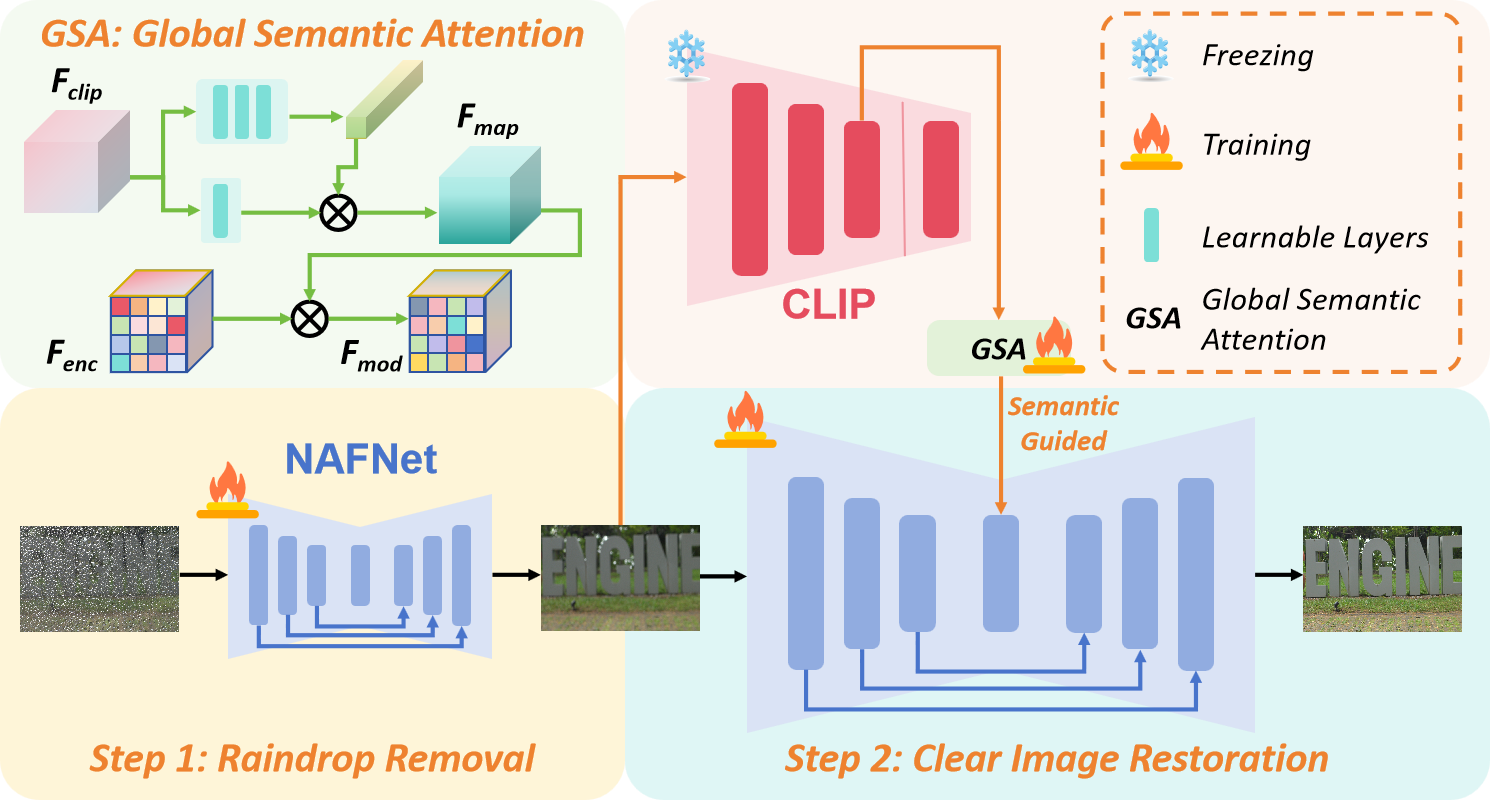}
    \caption{Architecture of the proposed GSA2Step framework by Team DGL\_DeRainDrop.}
    \label{fig:DGL_DeRainDrop}
\end{figure}

This team proposes a Global Semantic Attention (GSA) based two-step dual-focused image raindrop removal method (GSA2Step), as depicted in Fig.~\ref{fig:DGL_DeRainDrop}. They adopt a two-step approach where they break down the dual-focused image raindrop removal task into two sub-tasks: raindrop removal and clean image restoration. In the first step, they use a 32-width NAFNet~\cite{chen2022simpleNAFNet} to remove focused and defocused raindrops. In the second step, they employ a Global Semantic Attention (GSA) module that utilizes CLIP~\cite{radford2021learningclip} semantic features to guide a 64-width NAFNet in reconstructing the final clean image from the background obtained in step 1. The GSA mechanism extracts semantic features using the CLIP image encoder and effectively integrates them into the encoded features obtained from the NAFNet encoder to guide the subsequent decoding process. This architecture addresses various degradations including focused and defocused raindrops, as well as defocused backgrounds.

\noindent\textbf{Training Details.} They employ a multi-stage training approach to reduce the network's learning difficulty. Initially, they pretrain the 32-width NAFNet using the Drop and Blur subsets from the training dataset with only L1 loss. Subsequently, they freeze the pretrained NAFNet and train the entire framework using the Drop and Clear subsets, still with only L1 loss. Finally, they unfreeze the parameters of the first NAFNet and jointly fine-tune the entire framework using a combination of L1 loss, SSIM loss, and perceptual loss~\cite{johnson2016perceptual} with weight coefficients of 1.0, 0.5, and 0.01, respectively. The model is trained using the Adam~\cite{kingma2014adam} optimizer with learning rate initially set to $4\times e^{-4}$ and halved after every 200,000 iterations, with the final fine-tuning using a fixed learning rate of $1\times e^{-5}$. Training is conducted on a single NVIDIA RTX 4090 GPU with PyTorch~\cite{paszke2019pytorch} for approximately 4 days over 1000 epochs with no additional datasets utilized.

\subsection{xdu\_720}
\begin{figure}[htbp]
    \centering
    \includegraphics[width=1.0\linewidth]{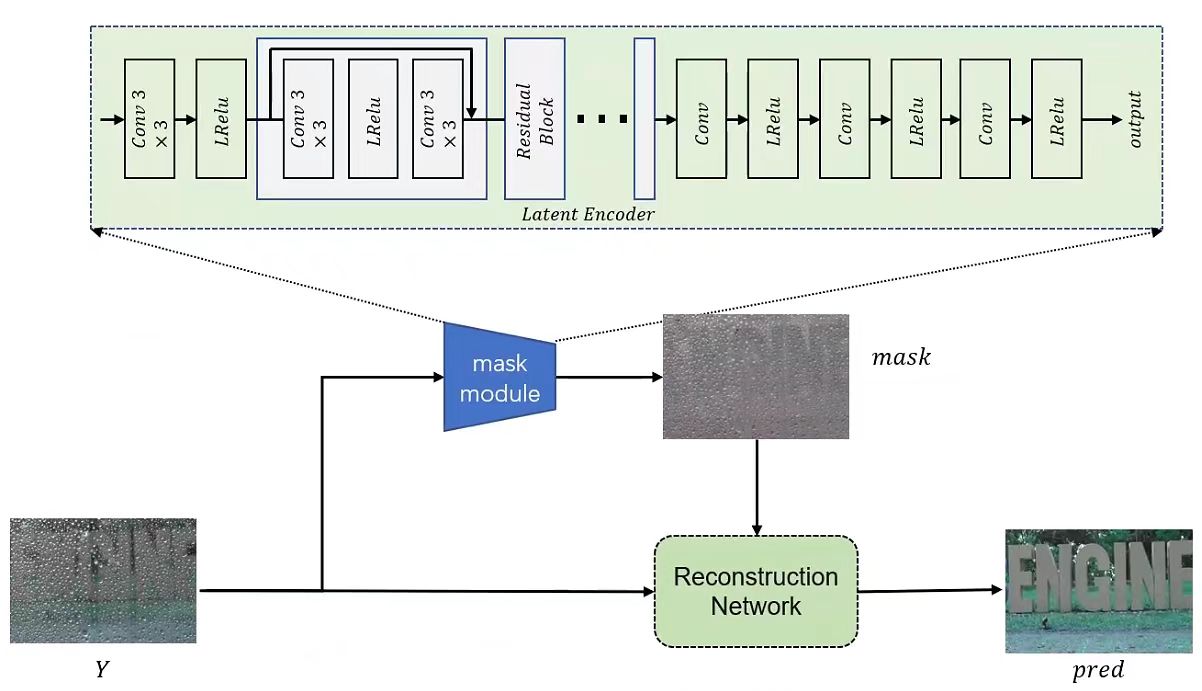}
    \caption{\yeying{The overall framework by Team xdu\_720.}}
    \label{fig:xdu_720}
\end{figure}

This team adopts a two-stage architecture combining a mask prediction network and a Transformer-based restoration model, as depicted in Fig.~\ref{fig:xdu_720}. In the first stage, a residual convolutional network is used to predict the raindrop mask from the input raindrop image, supervised by an L1 loss using ground truth masks obtained from the difference between the Raindrop and Blur images. In the second stage, the predicted mask is fused with features within a Transformer-based deraining network (FFTformer~\cite{fftformer}), where mask features are concatenated after each block. The final output is optimized with an MSE loss against the Clear image.

\noindent\textbf{Training Details.} Training is conducted in two stages. Stage one trains the mask prediction network for 20 epochs using L1 loss. Stage two trains the mask-prior-guided Transformer restoration network for 80 epochs using MSE loss. The predicted mask from stage one is used as additional guidance throughout the Transformer layers. No additional datasets or pre-training are used.

\noindent\textbf{Testing Details.} A subset of 3,000 image pairs from the training set is used for testing. During inference, the raindrop image is processed jointly by the mask prediction network and the FFTformer~\cite{fftformer}. The evaluation is based on PSNR, SSIM, and LPIPS metrics.

\subsection{EdgeClear-DNSST}
This team proposes EdgeClear-DNSST, a transformer-based encoder-decoder model designed for raindrop removal in dual-focused images. The model integrates Sparse-Sampling Attention for efficient long-range dependency modeling, combined with specialized modules including RaindropEdgeEnhancer, PyramidAttention, and RaindropFeatureModulation to address fine-grained raindrop degradation. The network adopts a multi-stage structure~\cite{zamir2022restormer} with four encoder and decoder stages, enhanced by skip connections and multi-scale feature fusion.

\noindent\textbf{Training Details.} The model is trained from scratch using only the day and night images from the RaindropClarity dataset~\cite{jin2024raindrop}, split into a 9:1 training-validation ratio. The training process utilizes the Adam optimizer with an initial learning rate of 0.0006, scheduled by cosine annealing (T\textsubscript{max}=100, $\eta_{\text{min}}$=$1\times e^{-6}$). The batch size is 24 per GPU with gradient accumulation (accumulate grad batches=2). Data augmentation includes random horizontal flips and $90^{\circ}$ rotations. A combination of Y-channel PSNR loss, SSIM loss, LPIPS perceptual loss, and difference-weighted loss is used. An early stopping strategy with a patience of 20 is applied based on validation PSNR.

\noindent\textbf{Testing Details.} During inference, they employ a sliding window strategy with a window size of 128$\times$128 and a 32-pixel overlap to handle images of arbitrary resolution. A weighted averaging method is used at the window edges to suppress boundary artifacts. Bilinear interpolation with aligned corners is adopted to mitigate edge color anomalies.

\subsection{MPLNet}
This team builds upon the multi-stage fusion network MPRNet~\cite{zamir2021mprnet} and enhances its performance under non-uniform illumination conditions by integrating a Global Illumination Detection Module (IllumAdjust)~\cite{cao2020global,wei2018deep}. IllumAdjust generates a single-channel illumination map through a series of convolution and attention layers, which is then embedded into the encoder-decoder framework via a custom-designed UniMetaFormer unit. The UniMetaFormer module, based on Metaformer~\cite{zhang2023metatransformer}, replaces the CAB modules in MPRNet’s encoder and decoder. It comprises a Dynamic Tanh normalization layer (DyT)~\cite{zhu2025dyt}, an Illumination-guided Channel Attention Block (ICAB), and a convolution-based MLP. ICAB integrates channel and spatial attention mechanisms guided by the illumination map, enhancing the model's adaptability to illumination variations. The network retains MPRNet’s three-stage progressive restoration and employs Supervised Attention Modules (SAM) for stage-wise feature fusion. An Original Resolution Block (ORB) is used in the final stage to further refine the output.

\noindent\textbf{Training Details.}
This team trained their model using the same dataset as in ~\cite{zamir2021mprnet}. The loss function combines Charbonnier Loss, Edge Loss, and LPIPS Loss to balance pixel-level accuracy and perceptual quality. They performed validation after each epoch and saved model checkpoints based on the best PSNR and LPIPS scores, as well as at fixed intervals. Gradient clipping was applied to stabilize training. All experiments were conducted on an NVIDIA GPU.

\subsection{Singularity}
% Done!
\begin{figure}[htbp]
    \centering
    \includegraphics[width=1.0\linewidth]{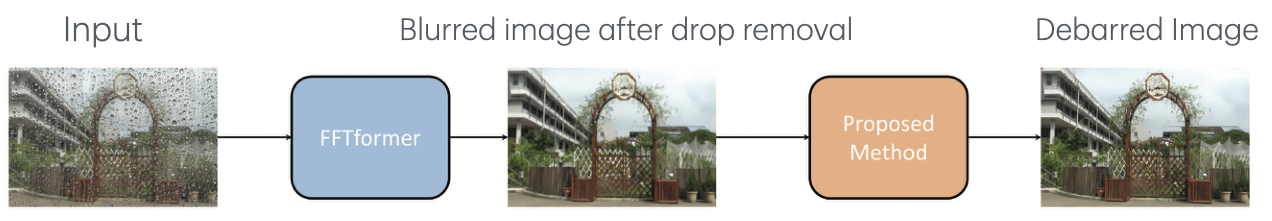}
    \caption{The overall pipeline proposed by Team Singularity.}
    \label{fig:Singularity_1}
\end{figure}
\begin{figure}
    \centering
    \includegraphics[width=1.0\linewidth]{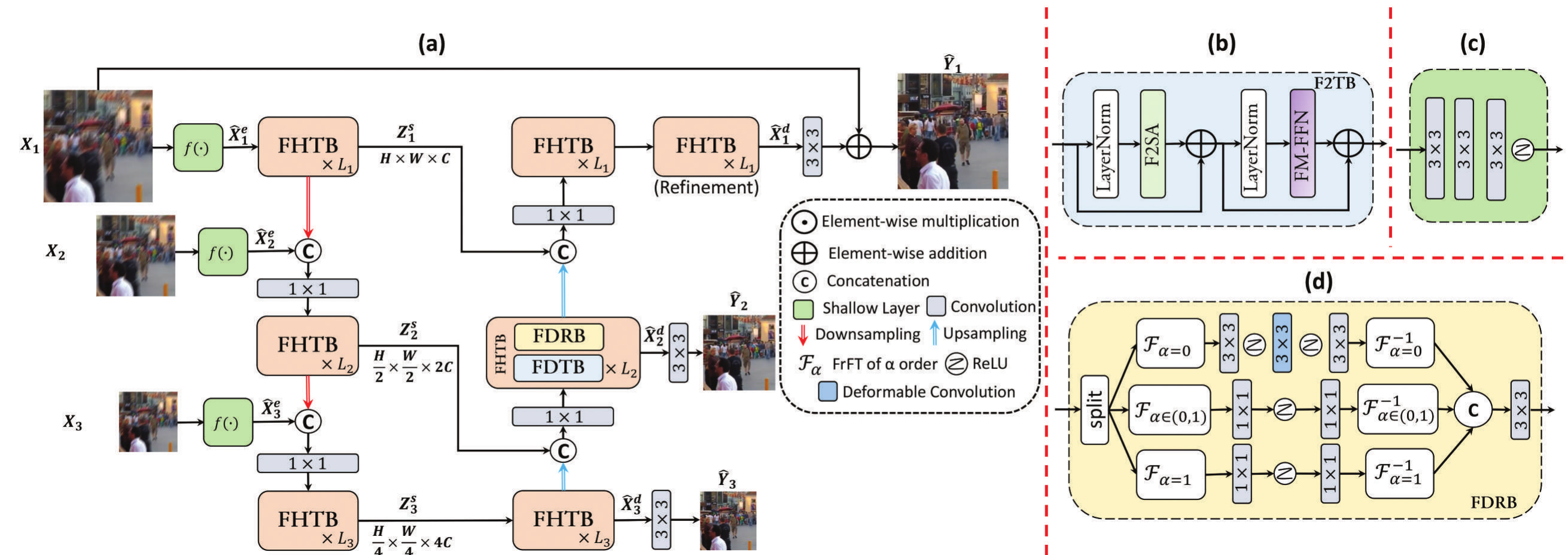}
    \caption{The overall architecture for deblurring proposed by Team Singularity.}
    \label{fig:Singularity_2}
\end{figure}
\begin{figure}
    \centering
    \includegraphics[width=1.0\linewidth]{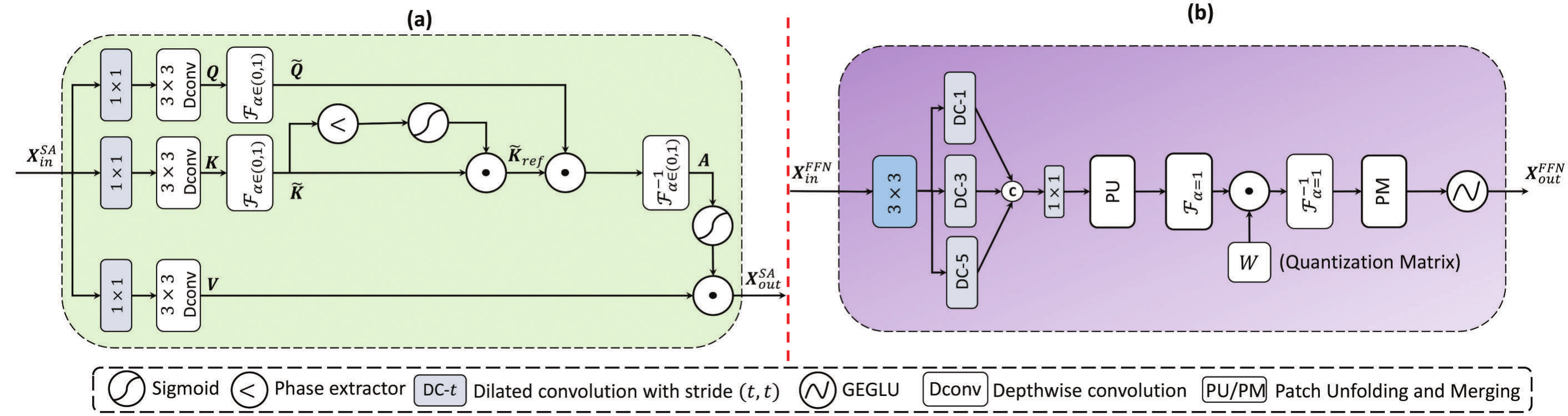}
    \caption{The (a) self-attention mechanism Fractional Frequency aware Self-Attention (F2SA), and the (b) Frequency quantized feed forward network FQ-FFN proposed by Team Singularity.}
    \label{fig:Singularity_3}
\end{figure}

This team proposes a two-step pipeline to address raindrop and blur removal challenges, as depicted in Fig.~\ref{fig:Singularity_1}. The approach leverages the distinct characteristics of raindrops and blurring artifacts, which require different handling strategies. By separating the tasks into two stages, they apply specialized techniques to effectively remove each type of artifact, leading to higher-quality image reconstruction. For raindrop removal, they utilize FFTformer~\cite{fftformer} to effectively handle the irregular patterns and occlusions caused by raindrops. For blur removal, they develop an enhanced version of F2former~\cite{paul2025f2former}, leveraging Fractional Fourier Transform (FRFT) for non-uniform deblurring, as shown in Fig.~\ref{fig:Singularity_2}. The blur removal framework consists of two components: FHTB and FDTB. Specifically, FHTB includes FDRB (similar to F3RB in \cite{paul2025f2former}) and FDTB, which use deformable convolutions to adapt to varying motion or blur patterns, improving edge restoration. FDTB uses an FRFT-based attention mechanism, inspired by~\cite{paul2025f2former}, and integrates quantization from FFTformer’s FSAS module to focus on relevant frequency components, ensuring efficient handling of the blur pattern at multiple levels. The details are illustrated in Fig.~\ref{fig:Singularity_3}.

\noindent\textbf{Training Details.} This team trains the raindrop and blur removal models separately. The two models are trained following the strategy of NAFNet~\cite{chen2022simple} and F2former~\cite{paul2024f2former}. They use the standard data augmentation, the Adam optimizer with default settings, and a learning rate of 1e-3. They update their model with a cosine annealing strategy for 600,000 iterations. Both models are trained on $256\times256$ images with a batch size of 8, and a patch size of $8\times8$ is used for self-attention. They apply L1 loss in both spatial and frequency domains, with adversarial training applied for 50,000 iterations to improve perceptual quality. All experiments are conducted on an 80GB Nvidia A100 GPU, with training configurations modified from NAFNet.

\subsection{VIPLAB}
% Done!
This team utilizes an efficient unified framework with a
two-stage training strategy to explore the weather-general and weather-specific features separation. The first training stage aims to learn the weather-general
features by taking the images under various weather conditions as inputs and generating the coarsely restored results. The second training stage aims to learn to adaptively expand the specific parameters for each weather type in the deep model, where the requisite positions for expanding weather-specific parameters are automatically learned. Finally, they adopt NAFNet~\cite{chen2022simpleNAFNet} to enhance the textures.

\noindent\textbf{Training Details.} They optimize the model for 200 epochs using AdamW optimizer with a learning rate of 1e-4. They employ data augmentation techniques, including random cropping and flipping.

\subsection{2077Agent}
% Done!
\begin{figure}[htbp]
    \centering
    \includegraphics[width=1.0\linewidth]{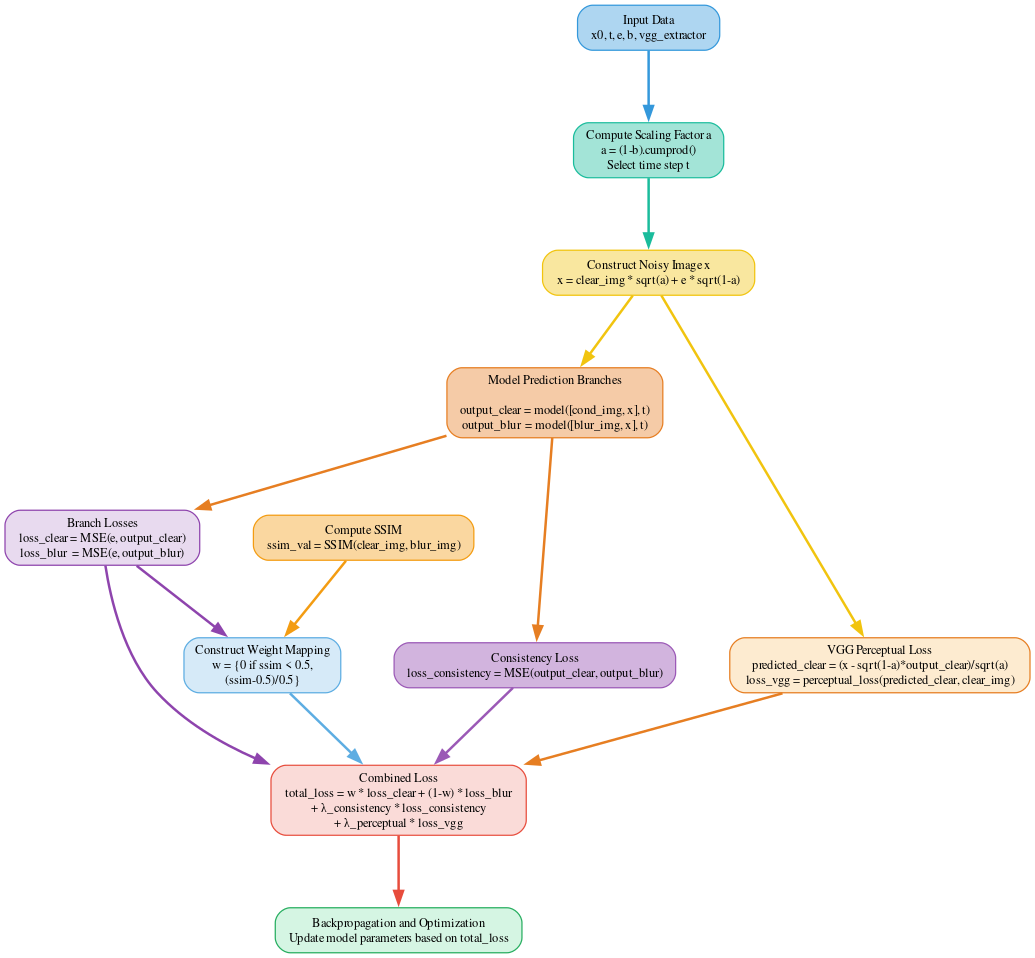}
    \caption{Architecture of the Loss Computation and Optimization Module by Team 2077Agent.}
    \label{fig:2077agent}
\end{figure}

This team builds the model based on the pre-trained DiT~\cite{peebles2023scalableDiT} model. They propose a cost-effective and efficient fine-tuning approach for dual-focused day-and-night raindrop removal by optimizing the loss function. Specifically, they introduce an adaptive loss function that integrates Structural Similarity (SSIM) and VGG-based perceptual losses. The SSIM-based weighting dynamically emphasizes regions with significant texture differences between clear and blurry images. Meanwhile, the perceptual loss leveraging VGG features ensures visually realistic outcomes. Additionally, a consistency constraint is incorporated to stabilize noise estimation between clear and blurred branches. This process is illustrated in Fig.~\ref{fig:2077agent}. They utilize additional datasets, including LHP-Rain~\cite{guo2023skylhp} and DIV2K~\cite{agustsson2017ntirediv2k}, for data augmentation to enhance the model’s generalization capability to unseen scenarios.

\noindent\textbf{Training Details.} They fine-tune the pre-trained DiT~\cite{peebles2023scalableDiT} model provided by the official repository using the Adam optimizer with an initial learning rate of $2\times e^{-5}$ for about 50 epochs. They resize the original training images to two scales (720$\times$480 and 360$\times$240) to enhance the generalization ability of the model towards diverse resolutions. They randomly crop the training patches into $64\times64$ for training. They optimize the DiT model with a linear beta scheduling strategy (ranging from 0.0001 to 0.02) with 1000 diffusion steps. Additionally, they develop a hybrid loss function with SSIM-based adaptive weighting, VGG perceptual loss, and consistency constraints, significantly improving the model’s detail recovery capability and visual quality of the restored images.

\noindent\textbf{Testing Details.} In the testing phase, they employed an implicit sampling method with 25 sampling steps to achieve efficient and high-quality raindrop removal. All output images are 720$\times$480 pixels. To mitigate boundary artifacts introduced during patch stitching, they utilize a grid-based overlapping strategy with a 16-pixel overlap during image reconstruction. Unlike the training phase, they do not crop patches during testing to ensure a comprehensive global evaluation of the full-resolution images. They conduct all evaluations with a single NVIDIA 4090 GPU.

\subsection{X-L}
% Done!
\begin{figure}[htbp]
    \centering
    \includegraphics[width=1.0\linewidth]{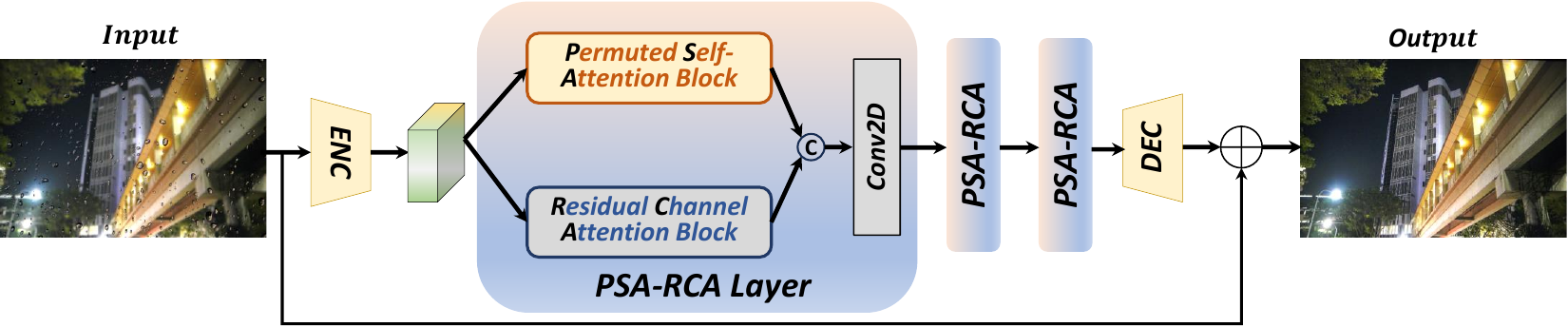}
    \caption{The overview of the proposed method by Team X-L.}
    \label{fig:X-L}
\end{figure}
This team develops the model following the popular encoder-decoder paradigm. They propose to enhance the middle features by a combination of Permuted Self-Attention blocks (PSA)~\cite{zhou2023srformerPSA} and Residual Channel Attention blocks (RCA)~\cite{zhang2018imageRCA}, as shown in Fig.~\ref{fig:X-L}. Each PSA-RCA layer comprises two parallel branches: one branch consists of residual channel attention blocks~\cite{zhang2018imageRCA}, which are designed to capture local features, while the other branch includes Permuted Self-Attention Blocks (PSAB)~\cite{zhou2023srformerPSA} that model global relationships with low computational complexity. This dual-branch architecture effectively integrates both local and global information, thereby enhancing the precision and efficiency of feature extraction. Following the processing through these three PSA-RCA layers, the refined features are passed to a feature decoder for additional processing and reconstruction, ultimately yielding a clear and detailed output.

\noindent\textbf{Training Details.} The model was trained using the provided dataset, with the L1 loss function employed to optimize the training process. No additional datasets were utilized during training. The training was conducted on a single NVIDIA 4090 GPU.

\noindent\textbf{Testing Details}
This team performed ensemble on the test data, adopting the same strategy as EDSR, which involves rotating the images at different angles to achieve this. This method enhances the robustness and quality of the final output by generating predictions from multiple perspectives. By integrating these predictions from various angles, the model can better capture the details and structural information in the images, significantly improving the performance of the image processing task.

\subsection{UIT-SHANKS}
% Done! with no latex file.
\begin{figure}[htbp]
    \centering
    \includegraphics[width=1.0\linewidth]{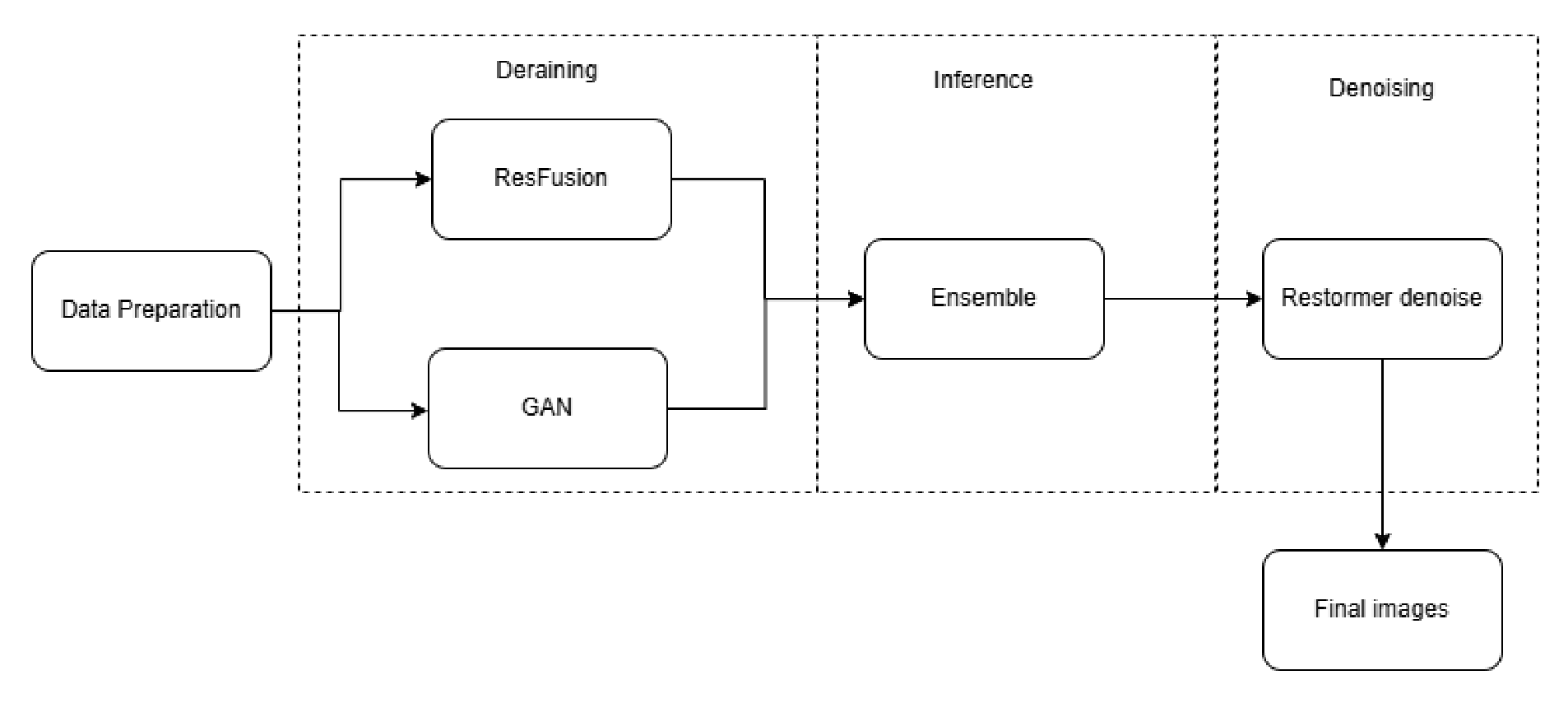}
    \caption{The proposed method of Team UIT-SHANKS.}
    \label{fig:UIT-SHANKS}
\end{figure}
This team proposes a two-stage approach to handle the challenge. In the first stage, they develop their model based on ReFusion~\cite{luo2023refusion}. In the second stage, they further optimize their model by incorporating the GAN-based training strategy, with a semi-supervised dataset derived from test data. They generate the dataset after training the model after 100 epochs. They build the discriminator in the GAN training phase based on UNet~\cite{unet} architecture. The overall framework is depicted in Fig.~\ref{fig:UIT-SHANKS}.

\noindent\textbf{Training Details.} In the first stage, they adopt the AdamW optimizer with the CosineAnnealing scheduler to optimize the model. They use MSE loss to obtain a fidelity-oriented model. In the second stage, they adopt the GAN-based training strategy with the standard GAN loss and VGG perceptual loss. Additionally, they leverage multiscale loss, SSIM loss and MASK loss to optimize the model. They finish their training with two T4 GPUs and a P100 GPU from the Kaggle platform.

\subsection{One Go Go}
% Done!
This team presents a two-stage approach for day and night raindrop removal for dual-focused images. Their method processes raindrop removal and defocus-blurring separately. They employ a retrained Restormer~\cite{zamir2022restormer} model specifically for raindrop removal in the first stage, while utilizing a fine-tuned IPT~\cite{chen2021pre} model to address the defocus-blurring issues present in the second stage. They maintain the same hyperparameters as in the original Restormer paper, with the number of channels set to [48, 96, 192, 384], respectively. For the implementation of IPT, they only retrain its head and tail components while preserving the core transformer blocks fixed. 

\noindent\textbf{Training Details.} They apply the same training process for both day and night images. However, they split the dataset into two functional subsets: one containing raindrops and one for blur correction. They develop two separate models rather than pursuing an end-to-end solution. The Restormer~\cite{zamir2022restormer} model was trained from scratch specifically for raindrop removal using the challenge dataset, while the IPT~\cite{chen2021pre} model leveraged pretrained weights with fine-tuning limited to only the head and tail components on the defocus-blurring subset. They optimize the model following the original Restormer approach, while adopting a patch size of $48\times48$ for IPT training. They accomplish their training with 4 NVIDIA GeForce RTX 3090 GPUs.

\noindent\textbf{Testing Details.} Their testing pipeline consists of two sequential processing stages. First, they apply the trained Restormer~\cite{zamir2022restormer} model to remove raindrops from the input images. Subsequently, they feed these processed images to the fine-tuned IPT~\cite{chen2021pre} model to restore the defocus-blurring. The inference costs approximately 455ms on a single GPU with their proposed testing pipeline.

\subsection{DualBranchDerainNet}
% Done!
\begin{figure}[htbp]
    \centering
    \includegraphics[width=1.0\linewidth]{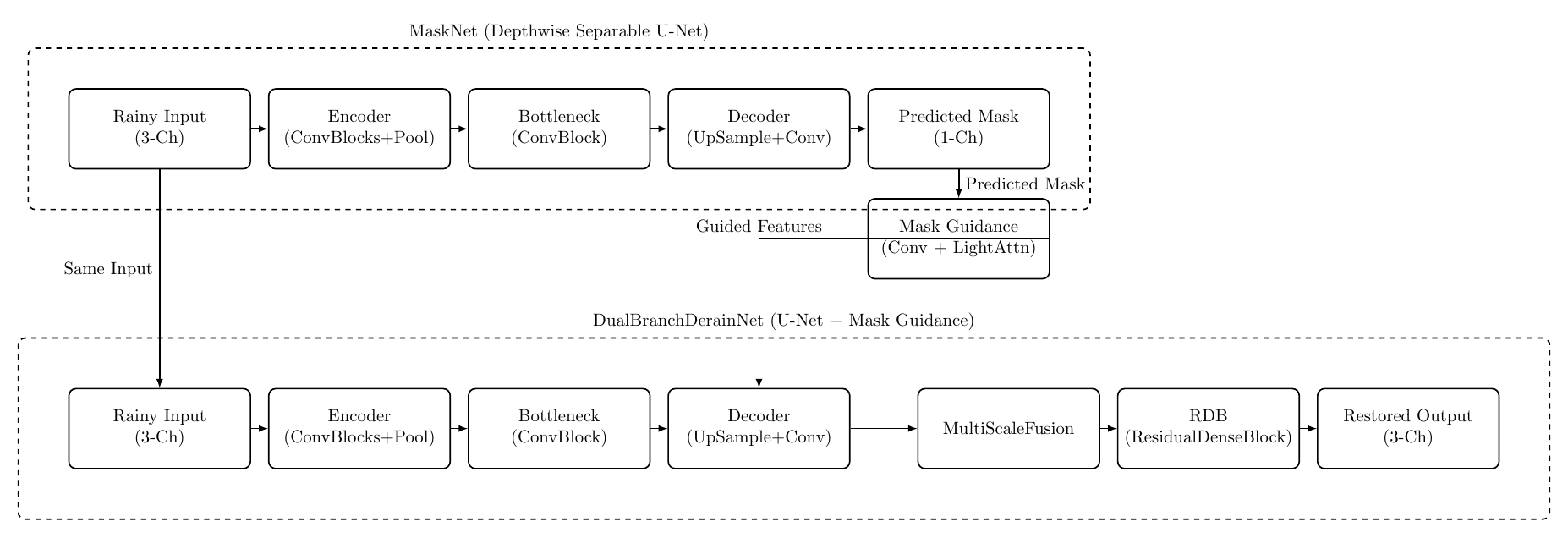}
    \caption{The network architecture proposed by Team DualBranchDerainNet. Zoom in for a better view.}
    \label{fig:DualBranchDerainNet}
\end{figure}
This team proposes a two-stage framework comprising a mask estimation network (MaskNet) and a dual-branch deraining network (DualBranchDerainNet), as shown in Fig.~\ref{fig:DualBranchDerainNet}. The MaskNet predicts rain masks by learning from the difference between the ``rain-focused'' and ``blur-focused'' inputs. The DualBranchDerainNet takes both the input image and the predicted mask, using a lightweight U-Net with multi-scale fusion and channel attention techniques to remove raindrops and restore clear details. They also adopt a WGAN-GP-based discriminator to improve the perceptual quality of the restored results, along with frequency and gradient losses for more faithful restoration.

\noindent\textbf{Training Details.} They train the proposed model for 300 epochs using Adam optimizer, with a batch size of 16 and an initial learning rate of 4e-3 on an A100 PCIE 40GB GPU. They adopt a combination of losses to optimize their model, including L1, SSIM, WGAN-GP, frequency domain, and gradient-based losses.
The approximate training time to obtain their model is 60 hours.

\noindent\textbf{Testing Details.} They evaluate their model by directly inferring the full-resolution images. The final output is a blend of the original input and a learned residual, guided by the predicted raindrop mask.\\

\subsection{QWE}
% Done!
This team develops their method upon the previous backbone, TransWeather~\cite{valanarasu2022transweather}, by leveraging prior knowledge from the pre-trained model and applying continuous learning on the competition dataset. The model architecture, algorithms, and module structure remain consistent with those outlined in the original TransWeather paper. However, they preprocess the dataset to align with the specific scenarios of this competition and introduce modifications to the loss function and training process.

\noindent\textbf{Training Details.}
They follow the officially released code of TransWeather~\cite{valanarasu2022transweather} and fine-tune the model for 200 epochs with RTX3050 GPUs.

\section*{Acknowledgments}
This work was partially supported by NSFC under Grant 623B2098 and the China Postdoctoral Science Foundation-Anhui Joint Support Program under Grant Number 2024T017AH. We thank the challenge sponsor: Eastern Institute for Advanced Study, Ningbo. This work was partially supported by the Humboldt Foundation. We thank the NTIRE 2025 sponsors: ByteDance, Meituan, Kuaishou, and University of Wurzburg (Computer Vision Lab).

%% file: sec/5_Appendix.tex
\appendix

\subsection*{Organizers}
\label{organizers}
\noindent\textit{\textbf{Title:}} NTIRE 2025 Challenge on Day and Night Raindrop Removal for Dual-Focused Images

\noindent\textit{\textbf{Members:}} \\
Xin Li\textsuperscript{1} (\textcolor{magenta}{xin.li@ustc.edu.cn}), \\
Yeying Jin\textsuperscript{2,3} (\textcolor{magenta}{jinyeying@u.nus.edu}),\\
Xin Jin\textsuperscript{4} (\textcolor{magenta}{jinxin@eitech.edu.cn}), \\
Zongwei Wu\textsuperscript{5} (\textcolor{magenta}{zongwei.wu@uni-wuerzburg.de}), \\Bingchen Li\textsuperscript{1} (\textcolor{magenta}{lbc31415926@mail.ustc.edu.cn}), \\ Yufei Wang\textsuperscript{6} (\textcolor{magenta}{ywang25@snap.com}), \\ 
Wenhan Yang\textsuperscript{7} (\textcolor{magenta}{yangwh@pcl.ac.cn}), \\
Yu Li\textsuperscript{8} (\textcolor{magenta}{liyu@idea.edu.cn}), \\
Zhibo Chen\textsuperscript{1} (\textcolor{magenta}{chenzhibo@ustc.edu.cn}), \\
Bihan Wen\textsuperscript{9} (\textcolor{magenta}{bihan.wen@ntu.edu.sg}), \\
Robby T. Tan\textsuperscript{2} (\textcolor{magenta}{robby.tan@nus.edu.sg}), \\
Radu Timofte\textsuperscript{5} (\textcolor{magenta}{Radu.Timofte@uni-wuerzburg.de})

\noindent\textit{\textbf{Affiliations:}}

\noindent\textsuperscript{1} University of Science and Technology of China

\noindent\textsuperscript{2} National University of Singapore

\noindent\textsuperscript{3} Tencent 

\noindent\textsuperscript{4} Eastern Institute of Technology, Ningbo

\noindent\textsuperscript{5} Computer Vision Lab, University of Würzburg

\noindent\textsuperscript{6} Snap Research

\noindent\textsuperscript{7} Pengcheng Laboratory

\noindent\textsuperscript{8} IDEA

\noindent\textsuperscript{9} Nanyang Technological University

\subsection*{Miracle}

\noindent\textit{\textbf{Title:}}  Semantics-Guided Two-Stage Raindrop Removal Network

\noindent\textit{\textbf{Members:}} Qiyu Rong (\textcolor{magenta}{20231083510916@buu.edu.cn}), Hongyuan Jing, Mengmeng Zhang, Jinglong Li, Xiangyu Lu, Yi Ren,
Yuting Liu and Meng Zhang

\noindent\textit{\textbf{Affiliations:}}

\noindent Beijing Union University

\subsection*{EntroVision}

\noindent\textit{\textbf{Title:}} Two-stage Multi-scale Transformer for Day and Night Raindrop Removal

\noindent\textit{\textbf{Members:}} Xiang Chen\textsuperscript{1} (\textcolor{magenta}{chenxiang@njust.edu.cn}), Qiyuan Guan\textsuperscript{2}, Jiangxin Dong\textsuperscript{1}, Jinshan Pan\textsuperscript{1}

\noindent\textit{\textbf{Affiliations:}}

\noindent\textsuperscript{1}Nanjing University of Science and Technology

\noindent\textsuperscript{2}Dalian Polytechnic University

\subsection*{IIRLab}
\noindent\textit{\textbf{Title:}} A raindrop removal method based on Histoformer

\noindent\textit{\textbf{Members:}} Conglin Gou\textsuperscript{1} (\textcolor{magenta}{gou\_conglin@tju.edu.cn}), Qirui Yang\textsuperscript{1}, Fangpu Zhang\textsuperscript{1}, Yunlong Lin\textsuperscript{2}, Sixiang Chen\textsuperscript{3}, Guoxi Huang\textsuperscript{4}, Ruirui Lin\textsuperscript{4},
Yan Zhang\textsuperscript{5}, 
Jingyu Yang\textsuperscript{1}, Huanjing Yue\textsuperscript{1}

\noindent\textit{\textbf{Affiliations:}}

\noindent \textsuperscript{1}TianJin University

\noindent \textsuperscript{2}Xiamen University

\noindent \textsuperscript{3}The Hong Kong University of Science and Technology, Guangzhou

\noindent \textsuperscript{4}University of Bristol, UK

\noindent \textsuperscript{5}National University of Singapore, Singapore

\subsection*{PolyRain}

\noindent\textit{\textbf{Title:}} Finetuning Dense X-Restormer for Image Deraining

\noindent\textit{\textbf{Members:}} Jiyuan Chen\textsuperscript{1} (\textcolor{magenta}{jiyuan.chen@connect.polyu.hk}), Qiaosi Yi\textsuperscript{1} (\textcolor{magenta}{qiaosi.yi@connect.polyu.hk}), Hongjun Wang\textsuperscript{2}, Chenxi Xie\textsuperscript{1}, Shuai Li\textsuperscript{1}, Yuhui Wu\textsuperscript{1}

\noindent\textit{\textbf{Affiliations:}}

\noindent\textsuperscript{1}The Hong Kong Polytechnic University

\noindent\textsuperscript{2}The University of Tokyo

%H3FC2Zbv   
\subsection*{H3FC2Z}

\noindent\textit{\textbf{Title:}} Dual kmeans fusion for the RainDrop task

\noindent\textit{\textbf{Members:}} Kaiyi Ma, Jiakui Hu (\textcolor{magenta}{jiakuihu29@gmail.com})

\noindent\textit{\textbf{Affiliations:}}

\noindent  Xidian University

% IIC Lab
\subsection*{IIC Lab}

\noindent\textit{\textbf{Title:}} FA-Mamba

\noindent\textit{\textbf{Members:}} Juncheng Li\textsuperscript{1} (\textcolor{magenta}{junchengli@shu.edu.cn}), Liwen Pan\textsuperscript{1}, Guangwei Gao\textsuperscript{2}

\noindent\textit{\textbf{Affiliations:}}

\noindent \textsuperscript{1}Shanghai University

\noindent \textsuperscript{2}Nanjing University of Posts and Telecommunications

% BUPT_CAT
\subsection*{BUPT CAT}

\noindent\textit{\textbf{Title:}} Consistent Patch Transformer for Dual-Focused Day and Night Raindrop Removal

\noindent\textit{\textbf{Members:}} Wenjie Li (\textcolor{magenta}{lewj2408@gmail.com}), ZhenyuJin, Heng Guo, Zhanyu Ma

\noindent\textit{\textbf{Affiliations:}}

\noindent Beijing University of Posts and Telecommunications

% WIRTeam
\subsection*{WIRTeam}
\noindent\textit{\textbf{Title:}} MPID: Image Deraining with Multi-Scale Prompt-Based Learning

\noindent\textit{\textbf{Members:}} Yubo Wang (\textcolor{magenta}{wangyubo@stu.hit.edu.cn}), Jinghua Wang

\noindent\textit{\textbf{Affiliations:}}

\noindent Harbin Institute of Technology (Shenzhen)

% GURain
\subsection*{GURain}
\noindent\textit{\textbf{Title:}} Unified Image Restoration for Rain and Blur Using Restormer with a Dynamic Rain-Aware Weighted $\mathcal{L}_1$ Loss

\noindent\textit{\textbf{Members:}} Wangzhi Xing  (\textcolor{magenta}{w.xing@griffith.edu.au}), Anjusree Karnavar, Diqi Chen, Mohammad Aminul Islam   

\noindent\textit{\textbf{Affiliations:}}

\noindent Griffith University

%BIT\_ssvgg
\subsection*{BIT\_ssvgg}

\noindent\textit{\textbf{Title:}} Hybrid Network of CNN and Transformer for Raindrop removal

\noindent\textit{\textbf{Members:}} Hao Yang (\textcolor{magenta}{3120235187@bit.edu.cn}), Ruikun Zhang, Liyuan Pan

\noindent\textit{\textbf{Affiliations:}}

\noindent Beijing Institute of Technology

% CisdiInfo-MFDehazNet
\subsection*{CisdiInfo-MFDehazNet}

\noindent\textit{\textbf{Title:}} MFDehazNet-An Easily Deployable Image Dehazing
Model for Industrial Sites

\noindent\textit{\textbf{Members:}} Qianhao Luo (\textcolor{magenta}{QianHao.Luo@cisdi.com.cn}), XinCao (\textcolor{magenta}{Xin.A.Cao@cisdi.com.cn})

\noindent\textit{\textbf{Affiliations:}}
CISDI Information Technology CO., LTD

\subsection*{McMaster-CV}

\noindent\textit{\textbf{Title:}} \textbf{RainHistoNet}: Single-Image Day and Night Raindrop Removal via Histogram-Guided Restoration

\noindent\textit{\textbf{Members:}} Han Zhou (\textcolor{magenta}{zhouh115@mcmaster.ca}), Yan Min, Wei Dong, Jun Chen

\noindent\textit{\textbf{Affiliations:}}

\noindent Department of Electrical and Computer Engineering, McMaster University

\subsection*{Falconi}

\noindent\textit{\textbf{Title:}} No Title

\noindent\textit{\textbf{Members:}} Taoyi Wu (\textcolor{magenta}{taoyiwu81@gmail.com}), Weijia Dou, Yu Wang, Shengjie Zhao

\noindent\textit{\textbf{Affiliations:}}

\noindent Tongji University

%Dfusion
\subsection*{Dfusion}

\noindent\textit{\textbf{Title:}} Dffusion: A new method to fuse existing solutions with simple CNN

\noindent\textit{\textbf{Members:}} Yongcheng Huang (\textcolor{magenta}{Y.Huang-51@student.tudelft.nl}), Xingyu Han (\textcolor{magenta}{X.Han-5@student.tudelft.nl}), Anyan Huang (\textcolor{magenta}{A.Huang-3@student.tudelft.nl})

\noindent\textit{\textbf{Affiliations:}}

\noindent Delft University of Technology

% RainMamba
\subsection*{RainMamba}

\noindent\textit{\textbf{Title:}} RainMamba: A Video Coarse-to-Fine Mamba for Video Raindrop Removal

\noindent\textit{\textbf{Members:}} Hongtao Wu\textsuperscript{1} (\textcolor{magenta}{wuhongtao@westlake.edu.cn}), Hong Wang\textsuperscript{2}, Yefeng Zheng\textsuperscript{1}

\noindent\textit{\textbf{Affiliations:}}

\noindent \textsuperscript{1}Medical Artificial Intelligence Laboratory, West Lake University

\noindent \textsuperscript{2}School of Life Science and Technology, Xi'an Jiaotong University

% RainDropX
\subsection*{RainDropX}

\noindent\textit{\textbf{Title:}} Leveraging Perceptual and Structural Constraints for Dual-Focused Raindrop Removal Using Restormer

\noindent\textit{\textbf{Members:}} Abhijeet Kumar (\textcolor{magenta}{ee23d406@smail.iitm.ac.in}), Aman Kumar, A.N. Rajagopalan

\noindent\textit{\textbf{Affiliations:}}

\noindent Indian Institute of Technology Madras

%Cidaut AI
\subsection*{Cidaut AI}

\noindent\textit{\textbf{Title:}} FracDeformer: Fractional Fourier Aware
Deformable Transformer for Day and Night Raindrop Removal

\noindent\textit{\textbf{Members:}} Marcos V. Conde (\textcolor{magenta}{marcos.conde@uni-wuerzburg.de}), Paula Garrido, Daniel Feijoo, Juan C. Benito

\noindent\textit{\textbf{Affiliations:}}
Cidaut AI

\section*{DGL\_DeRainDrop}

\noindent\textit{\textbf{Title:}} GSA2Step

\noindent\textit{\textbf{Members:}} Guanglu Dong (\textcolor{magenta}{dongguanglu@stu.scu.edu.cn}), Xin Lin, Siyuan Liu, Tianheng Zheng, Jiayu Zhong, Shouyi Wang, Xiangtai Li, Lanqing Guo, Lu Qi and Chao Ren

\noindent\textit{\textbf{Affiliations:}}

\noindent Sichuan University 

%xdu\_720
\subsection*{xdu\_720}

\noindent\textit{\textbf{Title:}}  Raindrop Mask Prior-Guided Deraining Transformer

\noindent\textit{\textbf{Members:}} Shuaibo Wang (\textcolor{magenta}{shbwang@stu.xidian.edu.c}), Shilong Zhang, Wanyu Zhou, Yunze Wu, Qinzhong Tan

\noindent\textit{\textbf{Affiliations:}}
Xi'dian University

\subsection*{EdgeClear-DNSST Team}

\noindent\textit{\textbf{Title:}} EdgeClear-DNSST: Edge-preserving Day-Night Sparse-Sampling Transformer

\noindent\textit{\textbf{Members:}} Jieyuan Pei (\textcolor{magenta}{peijieyuan@zjut.edu.cn}), Zhuoxuan Li

\noindent\textit{\textbf{Affiliations:}}

\noindent Zhejiang University of Technology \& Tongji University

%MPLNet
\subsection*{MPLNet}

\noindent\textit{\textbf{Title:}} MPLNet: Multi-Stage Progressive Learning with Illumination-Conscious Dynamic Transformer Networks

\noindent\textit{\textbf{Members:}} Jiayu Wang (\textcolor{magenta}{2024090902024@std.uestc.edu.cn)}), Haoyu Bian, Haoran Sun

\noindent\textit{\textbf{Affiliations:}}

\noindent University of Electronic Science and Technology of China

%Singularity
\subsection*{Singularity}

\noindent\textit{\textbf{Title:}} FracDeformer: Fractional Fourier Aware Deformable Transformer for Day and Night Raindrop Removal

\noindent\textit{\textbf{Members:}} Subhajit Paul (\textcolor{magenta}{subhajitpaul@sac.isro.gov.in})

\noindent\textit{\textbf{Affiliations:}}

\noindent Space Applications Centre (SAC)

\subsection*{VIPLAB}

\noindent\textit{\textbf{Title:}} Staged restoration: deblurring first, then detail enhancement

\noindent\textit{\textbf{Members:}} Ni Tang (\textcolor{magenta}{23020240157683@stu.xmu.edu.cn}), Junhao Huang, Zihan
Cheng, Hongyun Zhu, Yuehan Wu

\noindent\textit{\textbf{Affiliations:}}

\noindent School of Informatics, Xiamen University

\subsection*{2077Agent}

\noindent\textit{\textbf{Title:}} DiT‑RainRemoval: Fine‑Tuning a Pre‑Trained DiT Model for Dual‑Focus Raindrop Removal

\noindent\textit{\textbf{Members:}} Kaixin Deng\textsuperscript{1} (\textcolor{magenta}{kaixin.deng.t0@elms.hokudai.ac.jp}), Hang Ouyang\textsuperscript{2}, Tianxin Xiao\textsuperscript{2}, Fan Yang\textsuperscript{2}, Zhizun Luo\textsuperscript{2}

\noindent\textit{\textbf{Affiliations:}}

\noindent\textsuperscript{1}Hokkaido University

\noindent\textsuperscript{2}Chengdu University of Technology

\subsection*{X-L}

\noindent\textit{\textbf{Title:}} Permuted Self-Attention Network for Image Raindrop Removal

\noindent\textit{\textbf{Members:}} Zeyu Xiao\textsuperscript{1} (\textcolor{magenta}{zeyuxiao1997@163.com}), Zhuoyuan Li\textsuperscript{2}

\noindent\textit{\textbf{Affiliations:}}

\noindent\textsuperscript{1} National University of Singapore

\noindent\textsuperscript{2} University of Science and Technology of China

\subsection*{UIT-SHANKS}

\noindent\textit{\textbf{Title:}} Effective Raindrop Removal: An Integrated Deep Ensemble Approach with Semi-Supervised Learning and Restormer Denoising

\noindent\textit{\textbf{Members:}} Nguyen Pham Hoang Le (\textcolor{magenta}{22520982@gm.uit.edu.vn}), An Dinh Thien,  Son T. Luu, Kiet Van Nguyen

\noindent\textit{\textbf{Affiliations:}}

\noindent University of Information Technology, Vietnam National University, Ho Chi Minh City, Vietnam

\subsection*{One Go Go}

\noindent\textit{\textbf{Title:}} Deraining and defocus-blurring separately

\noindent\textit{\textbf{Members:}} Ronghua Xu (\textcolor{magenta}{202424131014T@stu.cqu.edu.cn}), Xianmin Tian, Weijian Zhou, Jiacheng Zhang

\noindent\textit{\textbf{Affiliations:}}

\noindent School of Big Data and Software Engineering, Chongqing University

\subsection*{DualBranchDerainNet}

\noindent\textit{\textbf{Title:}} A Lightweight Dual-Branch Raindrop Removal Method for Day and Night Scenes

\noindent\textit{\textbf{Members:}} Yuqian Chen (\textcolor{magenta}{213240247@seu.edu.cn})

\noindent\textit{\textbf{Affiliations:}}

\noindent Southeast University

\subsection*{QWE}

\noindent\textit{\textbf{Title:}} Optimization of raindrop Image Restoration Model Integrating Data Enhancement and Continuous Learning

\noindent\textit{\textbf{Members:}} Yihang Duan (\textcolor{magenta}{2432179908@qq.com}), Yujie Wu

\noindent\textit{\textbf{Affiliations:}}

\noindent Xidian University

\subsection*{Visual and Signal Information Processing Team}

\noindent\textit{\textbf{Title:}} UFormer-based Dual-Focus Raindrop Removal for NTIRE 2025

\noindent\textit{\textbf{Members:}} Suresh Raikwar (\textcolor{magenta}{suresh.raikwar@thapar.edu}), Arsh Garg, Kritika 

\noindent\textit{\textbf{Affiliations:}}

\noindent None

\subsection*{The Zheng family group}

\noindent\textit{\textbf{Title:}} No title

\noindent\textit{\textbf{Members:}} Jianhua Zheng (\textcolor{magenta}{460919144@qq.com}), Xiaoshan Ma, Ruolin Zhao, Yongyu Yang, Yongsheng Liang, Guiming Huang

\noindent\textit{\textbf{Affiliations:}}

\noindent Zhongkai University of Agriculture and Engineering

\subsection*{RainClear Pioneers}

\noindent\textit{\textbf{Title:}} RainClearNet: A Dual-Stream Physics-Guided Framework for Raindrop Removal

\noindent\textit{\textbf{Members:}} Qiang Li  (\textcolor{magenta}{li\_qiang@shu.edu.cn}), Hongbin Zhang, Xiangyu Zheng

\noindent\textit{\textbf{Affiliations:}}

\noindent Shanghai University